\documentclass[10pt,twocolumn,letterpaper]{article}

\usepackage{cvpr}
\usepackage{times}
\usepackage{epsfig}
\usepackage{graphicx}
\usepackage{amsmath}
\usepackage{amssymb}
\usepackage{bbm}
\usepackage[nocompress]{cite}
\usepackage{graphicx}
\usepackage{array, booktabs}
\usepackage[export]{adjustbox}
\usepackage{boldline} 
\usepackage{multirow}
\usepackage{placeins}

\usepackage{enumitem}

\setitemize[0]{leftmargin=10pt}

\newenvironment{tight_itemize}{
	\begin{itemize}[leftmargin=8pt]
		\setlength{\topsep}{0pt}
		\setlength{\itemsep}{0pt}
		\setlength{\parskip}{0pt}
		\setlength{\parsep}{0pt}
	}{\end{itemize}}

\usepackage[breaklinks=true,bookmarks=false]{hyperref}

\cvprfinalcopy 

\def\cvprPaperID{7254} 

\begin{document}

\title{MANTRA: Memory Augmented Networks for Multiple Trajectory Prediction}

\author{Francesco Marchetti ~~~ Federico Becattini ~~~ Lorenzo Seidenari ~~~ Alberto Del Bimbo\\
MICC, University of Florence\\
{\tt\small name.surname@unifi.it}
}

\maketitle
\thispagestyle{empty}

\begin{abstract}
   Autonomous vehicles are expected to drive in complex scenarios with several independent non cooperating agents. Path planning for safely navigating in such environments can not just rely on perceiving present location and motion of other agents. It requires instead to predict such variables in a far enough future. In this paper we address the problem of multimodal trajectory prediction exploiting a Memory Augmented Neural Network. Our method learns past and future trajectory embeddings using recurrent neural networks and exploits an associative external memory to store and retrieve such embeddings. Trajectory prediction is then performed by decoding in-memory future encodings conditioned with the observed past. We incorporate scene knowledge in the decoding state by learning a CNN on top of semantic scene maps.  Memory growth is limited by learning a writing controller based on the predictive capability of existing embeddings. We show that our method is able to natively perform multi-modal trajectory prediction obtaining state-of-the art results on three datasets. Moreover, thanks to the non-parametric nature of the memory module, we show how once trained our system can continuously improve by ingesting novel patterns.
\end{abstract}

\vspace{-10pt}
\section{Introduction}

\begin{figure}[t]
	\centering
	\includegraphics[width=\columnwidth]{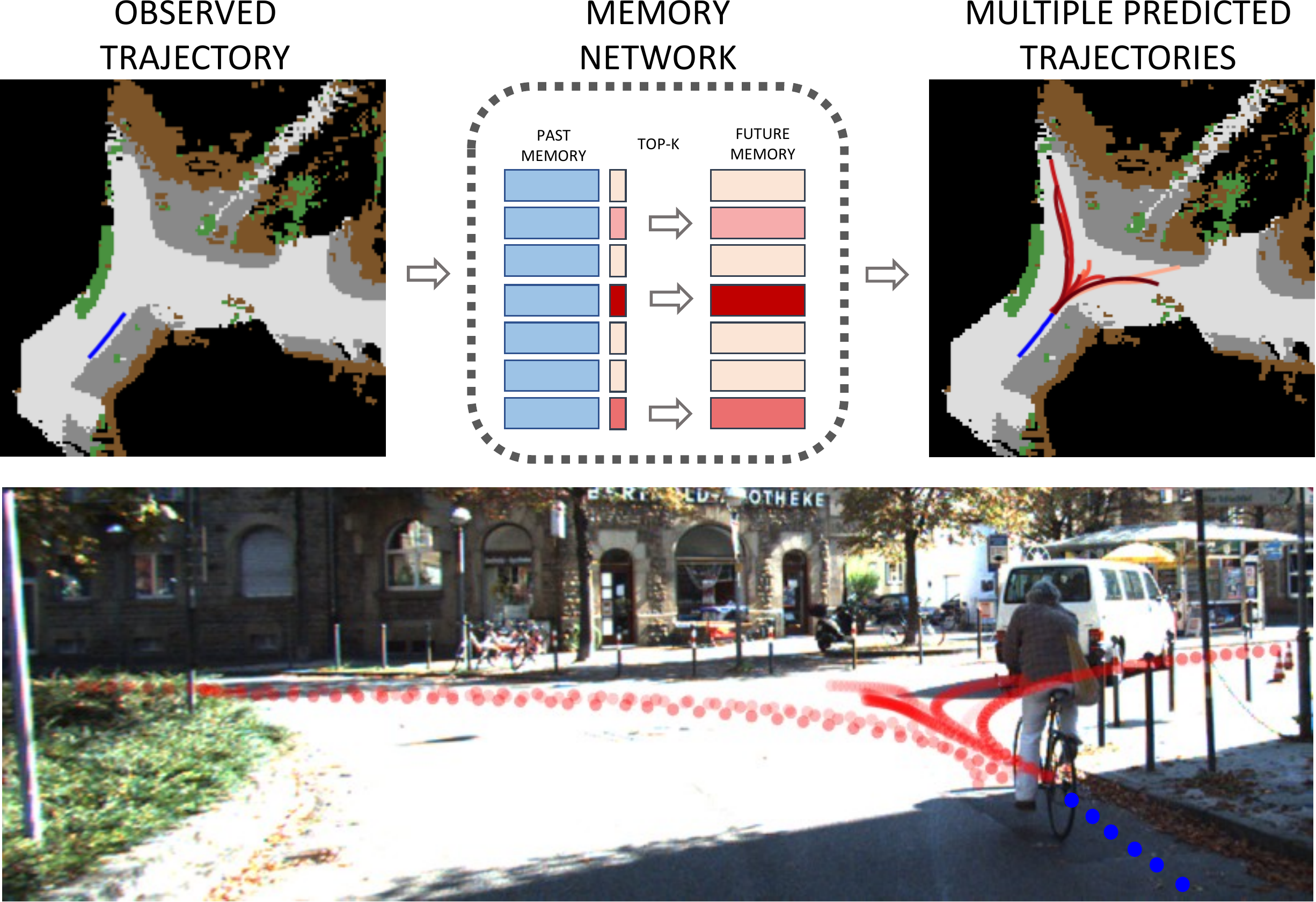}
	\caption{MANTRA addresses multimodal trajectory prediction. We obtain multiple future predictions given an observed past relying on a Memory Augmented Neural Network.}
	\label{fig:eyecatcher}
\end{figure}

What makes humans capable of succeeding in a large variety of tasks is the capacity to learn from experience, recalling past events and generalizing to new ones.
Learning to drive is a clear example of this ability. In recent years a lot of effort has been made to imitate this skill and to develop autonomous vehicles that are able to safely drive among other agents, either autonomous or driven by humans. 
Whereas remarkable progress has been made for automotive \cite{bojarski2016end, codevilla2018end, xu2017end}, current approaches still lack the ability to explicitly remember specific instances from experience when trying to infer possible future states of surrounding agents. This is particularly important for predicting future locations of moving agents, so to take appropriate decisions and avoid collisions or potentially dangerous situations. 
Predicting future trajectories of such agents is intrinsically multimodal: vehicle dynamics give rise to a set of similarly likely outcomes for an external observer (Fig. \ref{fig:eyecatcher}).

While humans can address this task by implicit learning, i.e. exploiting procedural memory (knowing how to do things) from similar scenarios of previous experience, without explicit and conscious awareness, for machines this task has proven to be extremely hard. Common machine learning models, such as Recurrent Neural Networks, fail to address it. They are capable to store past information into an internal state, updated at every time step, and make predictions based on long term patterns. But in such networks, memory is a single hidden representation and is only addressable as a whole. State to state transition is unstructured and global. Instead, an element-wise addressable memory would be useful to selectively access only relevant pieces of information. This would allow to peak into likely futures to guide predictions.

In this paper we present MANTRA: Memory Augmented Neural TRAjectory predictor. MANTRA is a novel approach implementing a persistent Memory Augmented Neural Network (MANN) for vehicle trajectory prediction.
In our model, an external associative memory is trained to write pairs of past and future trajectories and keep in memory only the most meaningful and non-redundant samples. 

The model incrementally creates a knowledge base that is used as experience to perform meaningful predictions. This mimics the way in which implicit human memory works. Since the knowledge base is built from trajectory samples, it can also include instances observed while the system is running, after it has been trained. In this way the system gains experience online increasing its accuracy and capability to generalise at no training cost.

To memorize samples, past and future trajectories are stored in the memory in an encoded form, separately. In fact, this permits to use the encoding of an observed trajectory as a memory key to read an encoded future and decode them jointly to generate a prediction. 
Therefore, the actual coordinates are obtained decoding a future read from memory, conditioning it with the observed past. In this way, the output is not a simple copy of previously seen examples, but is instead a newly generated trajectory obtained both from the system experience (i.e. its memory) and the instance observed so far. By reading multiple futures from memory, diverse meaningful predictions can be obtained. 
The main contributions of this paper are the following:
\vspace{-5pt}
\begin{tight_itemize}
	\setlength\itemsep{2px}
	\item We propose a novel architecture for multiple trajectory prediction based on Memory Augmented Neural Networks. To the best of our knowledge we are the first to adopt MANNs for trajectory prediction.
	\item Our formulation, exploiting an encoder-decoder pipeline augmented with an associative memory, is easier to inspect and provides naturally multimodal predictions, obtaining state-of-the-art results on three traffic datasets.
	\item Our model is able to improve incrementally, after it has been trained, when observing new examples online. This trait is important for industrial automotive applications and is currently lacking in other state of the art predictors.
\end{tight_itemize}


\section{Related Work}

\paragraph{Trajectory Prediction}
Significant effort has been made in the past years regarding trajectory prediction.
Several researchers have focused on trajectories of pedestrians~\cite{helbing1995social, pellegrini2009you, alahi2016social, gupta2018social, sadeghian2019sophie}, either regarded as individuals or crowds, also exploiting social behaviors and interactivity between individuals~\cite{helbing1995social, pellegrini2009you, alahi2016social,gupta2018social, lisotto2019social}.
While relevant for pedestrians, social behaviors are much less relevant for vehicles ~\cite{lee2017desire}. In this context, focus shifts instead on the observation of motion of the individual agents (their past trajectory) and the understanding of the surrounding environment ~\cite{lee2017desire, srikanth2019infer}. Traffic dynamics likely reduce to simpler scenarios where movement is limited and constrained by the environment. A notable exception is estimating lane changes on highways~\cite{kuefler2017imitating, deo2018multi}.
A few efforts have been made to understand and predict vehicle trajectories in urban scenarios~\cite{lee2017desire, srikanth2019infer, zyner2019naturalistic, ma2019trafficpredict}.
Among them, DESIRE~\cite{lee2017desire} uses a Variational Autoencoder for estimating a distribution from which future trajectories can be sampled. The method though is not able to generate confidence scores to provide a ranked set of trajectories. A large number of predictions is needed to cover all the search space and Inverse Optimal Control is then used to extract a final ranked subset. INFER~\cite{srikanth2019infer} instead exploits a fully convolutional model that takes into account intermediate semantic representations and generates multimodal heatmaps of possible future locations, then looking for peaks of the distribution.

In our work we address prediction of multiple vehicle trajectories in urban scenes. Examples of contexts where such multiple predictions may be necessary are roundabouts and crossroads where vehicles might take different equally possible paths. Differently from DESIRE~\cite{lee2017desire} our approach is able to directly estimate a small set of ranked trajectories which already exhibit sufficient diversity to cover multiple futures. Differently from INFER~\cite{srikanth2019infer} we directly work with coordinates instead of heatmaps, providing a better spatial resolution and more precise predictions.
Differently from both DESIRE and INFER, we train a Memory Augmented Neural Network model to generate multimodal trajectories, which to the best of our knowledge has never been used with this purpose. The usage of MANNs has two main advantages: (i) multiple futures can be read from memory for a given trajectory observation, making the model capable to predict multiple outcomes, complying to the multimodal nature of the problem; (ii) by retrieving a likely future from memory we can rely on an oracle that suggests what is going to happen in the near future.

A conceptually similar research direction to ours is the one of intention-based methods \cite{rhinehart2019precog, choi2019drogon, chai2019multipath}. Here, some anchor information (such as trajectories, actions or locations) are predefined and then used to guide predictions after estimating a probability distribution over each candidate.
In \cite{rhinehart2019precog}, predictions are conditioned by the state of a robot agent, for which a goal is given or estimated. The authors of \cite{choi2019drogon} propose a model for intersections that generates a likelihood over 5 fixed map zones, entailing different motion patterns. In \cite{chai2019multipath}, anchor trajectories are created with k-means and random sampling over training data.
To some extent, our memory entries can be interpreted as anchors encoding futures instead of intentions. However, we do not choose a reference agent to condition predictions or restrict the applicability to constrained scenarios.

In order to obtain meaningful predictions we also take context into account and its physical constraints. According to this, the set of trajectory proposals obtained by the MANN is refined by integrating knowledge of the surrounding environment using semantic maps. 
Finally, differently for prior work, our trajectory prediction model is also capable of growing online, improving incrementally its performance from new observations after it has been trained.

\begin{figure*}[ht]
	\centering
	\includegraphics[width=0.85\textwidth]{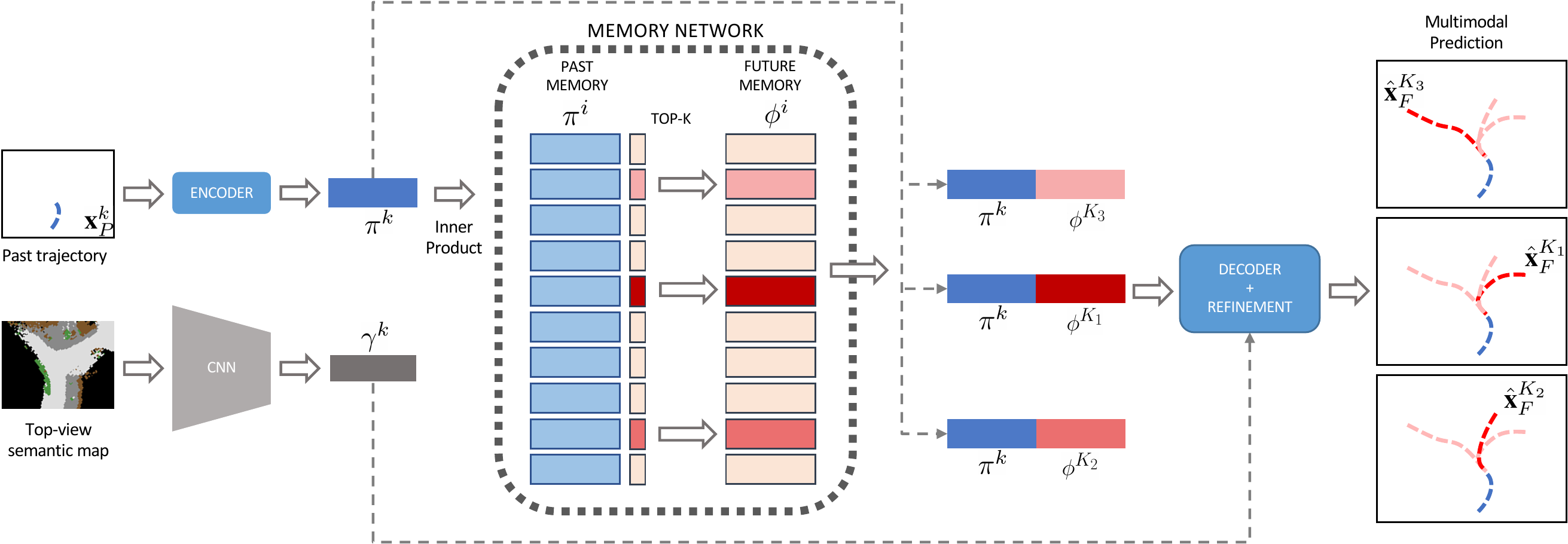}
	\caption{Architecture of MANTRA. The encoding of an observed past trajectory is used as key to read likely future encodings from memory. A multimodal prediction is obtained by decoding each future encoding, conditioned by the observed past. The surrounding context is processed by a CNN and fed to the Refinement Module to adjust predictions.}
	\label{fig:architecture}
\end{figure*}
\vspace{-10pt}
\paragraph{Memory Networks}
Neural networks with memory capabilities have been introduced to solve several machine learning problems which require to model a temporal dimension. The most common models are Recurrent Neural Networks (RNN) and their variants such as Long-Short Term Memories (LSTM)~\cite{hochreiter1997long} and Gated Recurrent Units (GRU)~\cite{cho2014learning}.
However, in these models memory is a single hidden state vector that encodes all the temporal information. So memory is addressable as a whole and they lack the ability to address individual elements of knowledge, necessary to apply algorithmic manipulation and rapid inference. Moreover, state to state transition is unstructured and global. Being the state updated at each time-step, eventually it fails to model very long term dependencies. Finally, the number of parameters is tied to the size of the hidden state. So adding knowledge from the external environment necessarily implies increasing the size of the state.
These characteristics prevent to use these models to effectively solve classes of problems like the one we address in this paper.

Recent works~\cite{kaiser2017learning, graves2014neural, santoro2016meta, sukhbaatar2015end, weston2014memory, rebuffi2017icarl, kumar2016ask, ma2018visual, yang2018learning, cai2018memory, vinyals2016matching, pritzel2017neural, xu2017few} have proposed Memory Augmented Neural Networks, or simply Memory Networks, to overcome the limitations of RNNs. The principal characteristic of this model is the usage of a controller network with an external element-wise addressable memory. This is used  to store explicit information and access selectively relevant items. The memory controller is trained to dynamically managed memory content optimizing predictions.  Differently from RNNs, state to state transitions are obtained through read/write operations and a set of independent states is maintained. An important consideration is that in Memory Networks the number of parameters is not tied to the size of the memory, i.e. increasing the memory slots will not increase the number of parameters. 

While introduced recently, a number of applications of this model have already appeared in literature. The first embodiment of a Memory Network was proposed in Neural Turing Machines (NTM)~\cite{graves2014neural} to perform algorithmic tasks, such as sorting or copying, which require sequential manipulation steps. Thanks to a fully differentiable controller, the model interacts with the memory through read/write operations. The architecture was later extended to perform one-shot learning in ~\cite{santoro2016meta}. Differently from NTM they trained the MANN to implement a Least Recently Used memory access strategy to write into rarely used locations.


In \cite{weston2014memory} MANNs have been proved to be able to effectively address Question Answering tasks, where the model has to answer questions related to a series of sentences. In \cite{sukhbaatar2015end} the same problem is solved with an End-to-End Memory Network with attention weights to shift importance from one sentence to another. Recent approaches have proposed a MANN to address the more complex problem of Visual Question Answering ~\cite{kumar2016ask, ma2018visual}, training the MANN to learn uncommon question-answer pairs.
Online learning has also been tackled using Memory Networks. Rebuffi \etal~\cite{rebuffi2017icarl} learn a classifier adding classes incrementally. MANNs for object tracking have been proposed where the model is trained to memorize templates, which are updated as the object is tracked ~\cite{yang2018learning}.

All these MANNs rely on episodic memories. The system learns to write and read from memory but the stored data is limited only to the current set of observations (such as a list of numbers to be sorted in ~\cite{graves2014neural} or a collection of sentences for question answering in ~\cite{weston2014memory}). 
Differently from prior work, we build a MANN with a memory that is not episodic. Instead, it acts like a persistent memory which stores an experience of relevant data to perform accurate predictions for any observation and not just for a restricted episode or set of samples.
The rationale behind this approach is that instead of solving simple algorithmic tasks as a Neural Turing Machine, we learn how to create a pool of samples to be used for future trajectory predictions.
The proposed model learns to store in memory only what is strictly needed to perform accurate predictions.  Our usage of MANN si close to~\cite{miller2016key}, but differs substantially. While they exploit the decoupling of embeddings to better fit data, we leverage the disjoint representation to create multiple outputs from a single input, leading to a fully multimodal predictive capability of the overall system. 


\section{Model}

We formulate the task of vehicle trajectory prediction as the problem of estimating $P(\hat{\textbf{x}}_F|\textbf{x}_P,\textbf{c})$, where $\hat{\textbf{x}}_F$ is the predicted future trajectory, $\textbf{x}_P$ is the observed trajectory (or \textit{past}) and \textbf{c} is a representation of the context (e.g. roads, sidewalks). We consider vehicle trajectories as a sequence of 2-dimensional spatial coordinates. The \textit{past} $\textbf{x}_P$ is given by its positions observed up to some reference point identified as \textit{present}. Similarly, the \textit{future} $\textbf{x}_F$ is the sequence of positions in which it will find itself at the next time steps.

\subsection{Memory Based Trajectory Prediction}

Given a sample trajectory $\textbf{x}^i = [\textbf{x}^i_P, \textbf{x}^i_F]$, let $\pi^i = \Pi(\textbf{x}^i_P)$ and $\phi^i = \Phi(\textbf{x}^i_F)$ be two encoding functions that map the 2D coordinates of past and future trajectories into two separate latent representations. Similarly, let $\Psi(\pi^i, \phi^i)$ be a function that decodes a pair of past-future encodings into the coordinates of the future sub-trajectory $\textbf{x}^i_F$, as shown in Fig.~\ref{fig:architecture}.

We define $M= \{ \pi^i, \phi^i \}$ as an associative key-value memory containing $|M|$ pairs of past-future encodings.
When a new trajectory $\textbf{x}_P^k$ is observed, its encoding $\pi^k$ is used as key to retrieve meaningful samples from memory. Note that observed trajectories are all considered to be \textit{past} trajectories, since the future counterpart is yet to be observed and is what we want to predict.
The memory addressing mechanism is implemented as a cosine distance between past encodings, which produces similarity scores $\{s_i\}$ over all memory locations:

\begin{equation}
s_i = \frac{\pi^k \pi^i }{\| \pi^k \| \| \pi^i \|} \quad  i=0,...,|M|
\end{equation}

According to the similarity scores, the future encodings of the top-K elements $\phi^j$ are separately combined with the encoding of the observed past $\pi^k$. The novel pairs of encodings are transformed into 2D coordinates using the decoding function $\Psi$: $\hat{\textbf{x}}^j_F = \Psi(\pi^k, \phi^j)$, with $j=1,...,K$.
Note that $\pi^k$ is fixed while $\phi^j$ varies depending on the sample read from memory.
Future encodings $\phi^j$ act as an oracle which suggests possible outcomes based on the past observation. This strategy allows the model to look ahead into likely futures in order to predict the correct one.
Since multiple $\phi^j$ can be used independently, we can decode multiple futures and obtain a multimodal prediction in case of uncertainty (e.g. a bifurcation in the road). 

\begin{figure}[t]
	\centering
	\includegraphics[width=0.95\columnwidth]{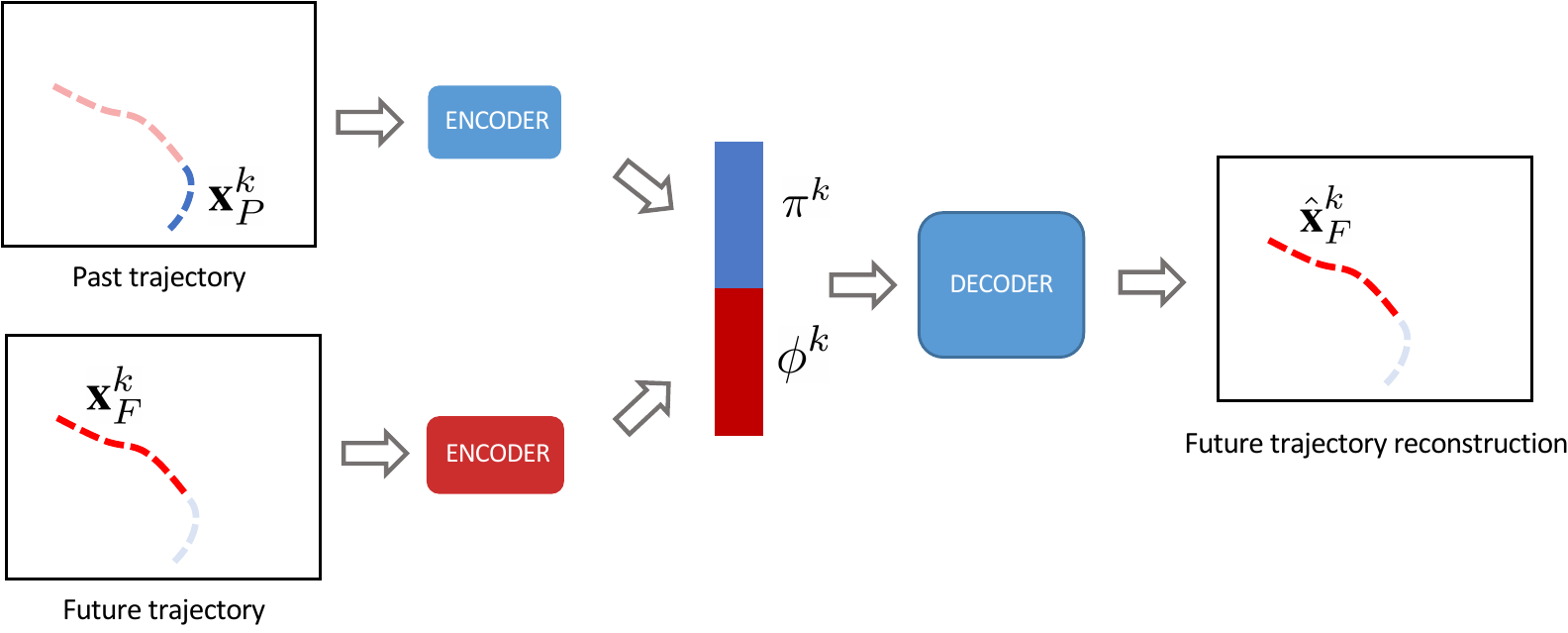}
	\caption{Representation learning: past and future trajectories are encoded separately; a decoder reconstructs future trajectory only.\label{fig:enc_dec}}
	\vspace{-10pt}
\end{figure}

\subsection{Feature Representation Learning}

The encoding-decoding functions $\Pi, \Phi, \Psi$ are trained jointly as an autoencoder, as shown in Fig.~\ref{fig:enc_dec}.
The encoders learn to map past and future points into a meaningful representation and the decoder learns to reproduce the future. Instead of using just the future as input, we condition the reconstruction process also with an encoding of the past. This is useful for two aspects. First, we are able to train two different encoders for past and future. The two encoders are used to obtain separate representations for both keys (past) and values (future) in memory. Second, we obtain reconstructions of the future that is compatible with the past. This is of crucial importance for prediction since at test time we synthesize trajectory encodings by combining past and future parts taken from different examples. This also allows to generate trajectories that differ from the ones in memory and are not just a simple copy of already observed samples.

\subsection{Memory controller}

Traditional Memory Augmented Neural Networks~\cite{graves2014neural, weston2014memory, sukhbaatar2015end} are designed to observe collections of data, usually referred to as episodes. The models are equipped with a working memory to store relevant information about the episode in order to generate a meaningful output for the episode. Yet memory is cleared for each episode and what is trained is the controller that decides what to read/write. The supervision for training stems from the cost function at the end of the episode, tracing gradients back to the controller.

As in standard memories, we train a controller to emit a write probability $P(w)$ every time that a sample is observed but, differently from these approaches, we train it to build a compact and expressive permanent memory.
Training such a controller might result challenging since $P(w)$ does not depend only on the intrinsic importance of the observed sample but also on the current state of the memory. To solve this issue, we do not rely on the prediction loss for supervision. We instead feed the reconstruction error $e$ to the controller, which decides if the network was sufficiently close to the ground truth. To enforce this behavior we define the controller loss $\mathcal{L}_c$ as:
\begin{equation}
\label{eq:controller_loss}
\mathcal{L}_c = e \cdot (1 - P(w)) + (1-e) \cdot P(w)
\end{equation}
where $e$ is assumed to have values in $[0, 1]$.
When the error is low, i.e. $e \rightarrow 0$, then
\begin{equation}
\mathcal{L}_c \approx P(w)
\end{equation}
therefore the write probability is minimized.

Conversely, when $e \rightarrow 1$, then
\begin{equation}
\mathcal{L}_c \approx 1 - P(w)
\end{equation}
and the controller maximizes the write probability.

What the controller is learning is an adaptive threshold on the reconstruction error, that allows to store in memory only what is useful to predict accurately, limiting redundancy. If the model exhibits a large prediction error, the controller writes the current sample with its ground truth future encoding in memory. When this happens, it indicates that the memory lacks samples to accurately reconstruct the future. Hence, by writing the sample in memory, the model will improve its prediction capabilities.

To satisfy the assumption of a bounded error function with values in $[0,1]$ for the controller loss of Eq. \ref{eq:controller_loss}, we introduce an adaptive miss rate error function with a threshold depending on the timestep:
\begin{equation}
e = 1 - \frac{1}{N}\sum_{i=1}^N \mathbbm{1}_i(\hat{\textbf{x}}_F, \textbf{x}_F)
\end{equation}

where $\mathbbm{1}_i(\hat{\textbf{x}}_F, \textbf{x}_F)$ is an indicator function equal to 1 if the $i$-th point of the prediction $\hat{\textbf{x}}_F$ lays within a threshold $th$ from the ground truth and 0 otherwise. We use a different threshold for each timestep, allowing a given uncertainty for the farthest point (4 seconds) and linearly decreasing towards 0. In our experiments we use $th_{4s}=2m$.



\subsection{Iterative Refinement Module}
To ensure compatibility with the environment, we refine predictions with an iterative procedure. Similarly to DESIRE \cite{lee2017desire}, we adopt a feature pooling strategy: first, a CNN extracts a feature map $\gamma^k$ from the context $\textbf{c}$; then, predictions are overlapped with the feature map and, for each time step coordinates, we extract the correspondent feature values (one per channel); finally, the resulting vector is fed to a GRU and a fully connected that output trajectory offsets.

The CNN is: $8\times(k3, s2, p1); 16\times(k3, s1, p1)$, where $k$ is kernel size, $s$ stride, $p$ padding. Both layers have BatchNorm and ReLU. The GRU has a hidden state size of 48.
We do 4 iterations and we observed that increasing them does not introduce substantial changes.

\subsection{Training}
\label{sec:training}
We train our model to observe 2 seconds trajectories and predict 4 seconds in the future. 
To achieve translation and rotation invariance, each trajectory is normalized by shifting the present in the origin and rotating the trajectory in order to make it  tangent with the Y-axis in the origin. In this way all futures start from (0, 0) in an upward direction.

First, a pretraining of both the encoders and the decoder is done jointly as an autoencoder. To do so, we feed pairs of past and future trajectories belonging to the same samples, reconstructing only the future coordinates.
We then train the memory controller, exploiting the learned past encoder and future decoder and resetting memory after every epoch. As controller we use a linear layer with sigmoid activation. The trained controller allows the memory to be filled with useful and non-redundant training samples by iterating over the training set and measuring their reconstruction error. While in principle the order in which samples are presented to the memory for writing may result in different final content, in our experiments we found that this does not affect the final prediction result. As a last step, we jointly train the refinement module and finetune the decoder. Here we feed the decoder with past and future encodings belonging to different samples, since the future is read from memory.


The two encoders and the decoder are implemented as Gated Recurrent Units with a 48-dimensional hidden state for each encoder and 96-dimensional for the decoder.
The GRU in the refinement module is initialized with the past embedding and takes as input the predicted coordinates. This provides the module with complete information about the whole trajectory.
We optimize $\mathcal{L}_c$ defined in Eq.~\ref{eq:controller_loss} to train the controller and a Mean Squared Error loss for decoder and refinement.
All components are trained with the Adam optimizer using a learning rate of 0.0001.

\vspace{-5px}
\section{Experiments}

\subsection{Datasets}
\label{sec:datasets}
\paragraph{KITTI~\cite{geiger2012we}}
The dataset includes many annotations such as Velodyne LiDAR 3D scans, object bounding boxes and tracks, calibration, depth and IMU. Not all data is present for every video so we used the ones categorized as \textit{KITTI Raw Data}, following the split of DESIRE~\cite{lee2017desire}. Although the split is known, how to divide trajectories in data chunks is not. To obtain samples we collect 6 seconds chunks (2 seconds for past and 4 for future) in a sliding window fashion from all trajectories in the dataset, including the ego-vehicle. We obtain 8613 top-view trajectories for training and 2907 for testing. Note that these numbers are different from the original DESIRE split since they claim to gather 2509 trajectories in total. To favor reproducibility and future comparison we will publicly release our version of the dataset upon publication. Since top-view maps are not provided by KITTI, we project semantic labels of static categories obtained with DeepLab-v3+~\cite{chen2017rethinking} from all frames in a common top-view map using the Velodyne 3D point cloud and IMU. The resulting maps have a spatial resolution of 0.5 meters, and will be released along with the trajectories.

Another smaller version of KITTI for trajectory prediction has been recently proposed by \cite{srikanth2019infer} and is publicly available. The authors propose 5 different train/test splits and average results over all runs, so we follow this evaluation protocol. We report experiments on both variants of KITTI. In the following we will refer to KITTI as our split obtained following DESIRE, unless stated otherwise.

\vspace{-5px}
\paragraph{Oxford RobotCar~\cite{maddern20171} \& Cityscapes~\cite{cordts2016cityscapes}}
The two datasets RobotCar and Cityscapes have been adapted for trajectory prediction in~\cite{srikanth2019infer} to show zero-shot transfer capabilities on different domains. Of particular interest is the ability to transfer to RobotCar since the sequences are acquired in the UK where cars drive on the left-side of the road. RobotCar has 6 seconds trajectories divided into 2 seconds for past and 4 for future. Cityscapes instead has shorter videos and predictions are made only up to one second in the future, as done in~\cite{srikanth2019infer}.

\subsection{Evaluation metrics and Baselines}
We report results in two common metrics for vehicle trajectory prediction: \textit{Average Displacement Error} (ADE) and \textit{Final Displacement Error} (FDE), where ADE is the average L2 error between all future timesteps and FDE (sometimes referred to as Horizon error) is the error at a given timestep. As in~\cite{lee2017desire, srikanth2019infer} we take the best out of $K$ predictions to account for the intrinsic multimodality of the task.
We compare our approach with several baselines: a linear coordinate regressor (\textit{Linear}); a Multi-Layer Perceptron with two layers trained as a coordinate regressor (\textit{MPL}); a Kalman filter~\cite{kalman1960new}, with a constant speed model used to propagate the estimate without incorporating measures (\textit{Kalman}).
We implemented and tested the baselines on the KITTI dataset to show comparable results. When available we also report existing baselines from the literature.

\begin{table}[t]
	\begin{center}
		\resizebox{0.7\textwidth}{!}{\begin{minipage}{\textwidth}
				\begin{tabular}{c|cccc|cccc}
					& \multicolumn{4}{c|}{ADE}       & \multicolumn{4}{c}{FDE}       \\
					Method & \multicolumn{1}{c}{1s} & \multicolumn{1}{c}{2s} & \multicolumn{1}{c}{3s} & \multicolumn{1}{c|}{4s} & \multicolumn{1}{c}{1s} & \multicolumn{1}{c}{2s} & \multicolumn{1}{c}{3s} & \multicolumn{1}{c}{4s} \\ \hline
					Kalman          &   0.51   &   1.14   &   1.99   &   3.03   &   0.97   &   2.54   &   4.71   &   7.41   \\
					Linear          &   0.20   &   0.49   &   0.96   &   1.64   &   0.40   &   1.18   &   2.56   &   4.73   \\
					MLP      &   0.20   &   0.49   &   0.93   &   1.53   &   0.40   &   1.17   &   2.39   &   4.12   \\
					
					MANTRA (top 1)  & 0.24      & 0.57   & 1.08    & 1.78   & 0.44      & 1.34      & 2.79      & 4.83   \\
					MANTRA (top 5)  & 0.17  & 0.36  & 0.61      &  0.94     & 0.30   & 0.75   & 1.43   & 2.48     \\
					MANTRA (top 10) & 0.16  & 0.30  &  0.48     &  0.73     & 0.26      &  0.59     & 1.07      &  1.88     \\
					MANTRA (top 20) & \textbf{0.16}  & \textbf{0.27}  & \textbf{0.40}      &  \textbf{0.59}     & \textbf{0.25}    & \textbf{0.49}    & \textbf{0.83}      &  \textbf{1.49}     \\ \hline \hline
					DESIRE (top 1)~\cite{lee2017desire}  &  -    &   -   &   -   &   -   &   0.51   &   1.44   &   2.76   &   4.45   \\
					DESIRE (top 5)~\cite{lee2017desire})  &   -   &   -   &   -   &   -   &   0.28   &   0.67   &   1.22   &   2.06   \\
					DESIRE (top 20)~\cite{lee2017desire})  &   -   &   -   &   -   &   -   &   -   &   -    &   -   &   2.04   \\ \hline                   
				\end{tabular}
		\end{minipage}}
		\caption{Results on the KITTI dataset. Results obtained by DESIRE are given as reference even if not comparable, due to the data collection process.}
		\label{tab:kitti}
		\vspace{-15pt}
	\end{center}
\end{table}

\subsection{Results}
Table~\ref{tab:kitti} shows the results on the KITTI dataset. Simply propagating the trajectory with a Kalman filter proves to be insufficient to accurately predict future positions, especially over long time spans, with a FDE@4s higher than 7m. 
Learning based baselines all perform better than Kalman filter, with the Multi-Layer Perceptron performing slightly better than the linear regressor.

Models that generate a single prediction fail to address the multimodality of the task, since they are trained to lower the error with a single output even when there might be multiple equally likely desired outcomes. What may happen is that in front of a bifurcation, the model predicts an average of the two possible trajectories, trying to satisfy both scenarios. Examples of this behavior are shown in Fig.~\ref{fig:qualitative}. Each prediction of MANTRA instead follows a specific path, ignoring the others. This leads to high errors on some examples when generating only one future, since the model may decide to follow a different likely path. On the other hand as soon as we generate $K$ multiple predictions, the top-$K$ error drastically decreases since we are able to cover diverse future paths.
We also report results from DESIRE~\cite{lee2017desire} varying $K$. Even though these results are not directly comparable as explained in Section~\ref{sec:datasets}, it is interesting to observe how DESIRE quickly saturates when increasing $K$, while our method keeps lowering the error significantly. This suggests that MANTRA samples a higher diversity of futures both at a coarse level (i.e. taking one road or another) and at a fine level (i.e. different behaviors on the same road).
Some qualitative results on KITTI are shown in Fig~\ref{fig:qualitative}, comparing them with the baselines.

Additionally, we evaluate MANTRA on the KITTI split proposed in~\cite{srikanth2019infer}, as shown in Table~\ref{tab:kitti-infer}. Here we also report some available baselines from the state of the art, both for single and multimodal predictions. With $K=1$ our method performs better or on par with INFER~\cite{srikanth2019infer} at low timesteps, yet losing some precision at 4s. Increasing K instead we are able to largely outperform INFER over all timesteps.

\begin{table}[t]
	\begin{center}
		\resizebox{0.7\textwidth}{!}{\begin{minipage}{\textwidth}
				\begin{tabular}{c|cccc|cccc}
					& \multicolumn{4}{c|}{ADE}       & \multicolumn{4}{c}{FDE}       \\
					Method & \multicolumn{1}{c}{1s} & \multicolumn{1}{c}{2s} & \multicolumn{1}{c}{3s} & \multicolumn{1}{c|}{4s} & \multicolumn{1}{c}{1s} & \multicolumn{1}{c}{2s} & \multicolumn{1}{c}{3s} & \multicolumn{1}{c}{4s} \\ \hline
					Kalman          &   0.33  &  0.54   &  0.93   & 1.4    &   0.46   &  1.18   &  2.18   &  3.32   \\
					Linear          &   0.31  &  0.56   &  0.89   &   1.28   &   0.47   &   1.13  &  1.94   &   2.87  \\
					MLP      &   0.30   &  0.54   &   0.88  &  1.28   &  0.46    &  1.12   &  1.94   &  2.88   \\
					RNN Enc-Dec ~\cite{virdi2017using}     &   0.68   &  1.94   &  3.20   &   4.46  &  -   &   -  &  -   &   -  \\
					Markov ~\cite{srikanth2019infer}   		&   0.70  &  1.41   &   2.12  &  2.99   &  -   &  -    &  -   &   -  \\ 
					
					Conv-LSTM (top 5) ~\cite{srikanth2019infer}  &   0.76  &  1.23   &  1.60   &  1.96   &  -   &  -   &  -   &   -  \\
					INFER (top 1)~\cite{srikanth2019infer}       &   0.75  &  0.95   &  1.13   &  1.42   &   1.01  &  1.26   &  1.76   &  2.67   \\
					INFER (top 5)~\cite{srikanth2019infer}       &   0.56  &  0.75   &  0.93   &  1.22   &  0.81   &  1.08   &  1.55   &  2.46   \\ 
					MANTRA (top 1)  &  0.37 & 0.67  & 1.07  & 1.55  &  0.60     &  1.33     & 2.32      & 3.50      \\ 
					MANTRA (top 5)  &  0.33 & 0.48  & 0.66      &  0.90     &   0.45      &  0.78     &  1.22   & 2.03  \\ 
					MANTRA (top 10) & 0.31   & 0.43   & 0.57    & 0.78   & 0.43      &  0.67     &  1.04     &  1.78     \\
					MANTRA (top 20) & \textbf{0.29}  & \textbf{0.41}  & \textbf{0.55}      &  \textbf{0.74}     &  \textbf{0.41}     & \textbf{0.64}      & \textbf{1.00}      &  \textbf{1.68}     \\ \hline                 
				\end{tabular}
		\end{minipage}}
		\caption{Results on the KITTI dataset (INFER split).}
		\label{tab:kitti-infer}
		\vspace{-15pt}
	\end{center}
\end{table}
\newcommand{\qualw}{.26\textwidth}
\begin{figure*}
	\centering
	\begin{tabular}{ccc}
		\includegraphics[width=\qualw, trim=50 105 50 90, clip]{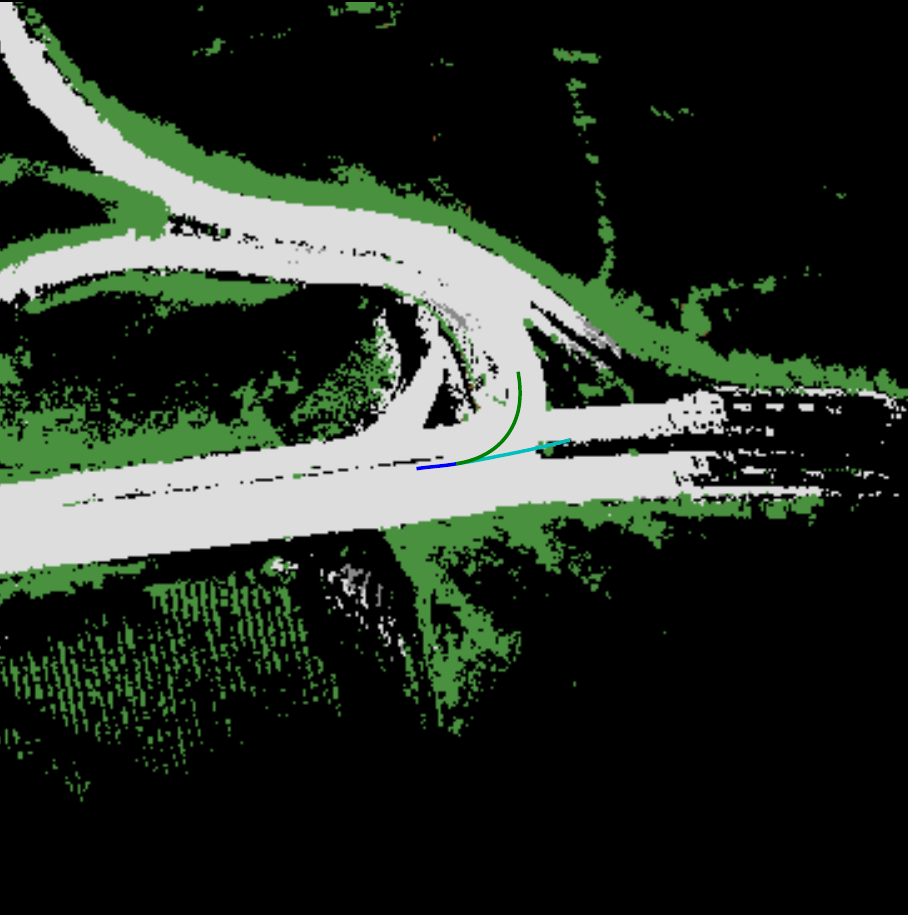} &
		\includegraphics[width=\qualw, trim=50 105 50 88.5, clip]{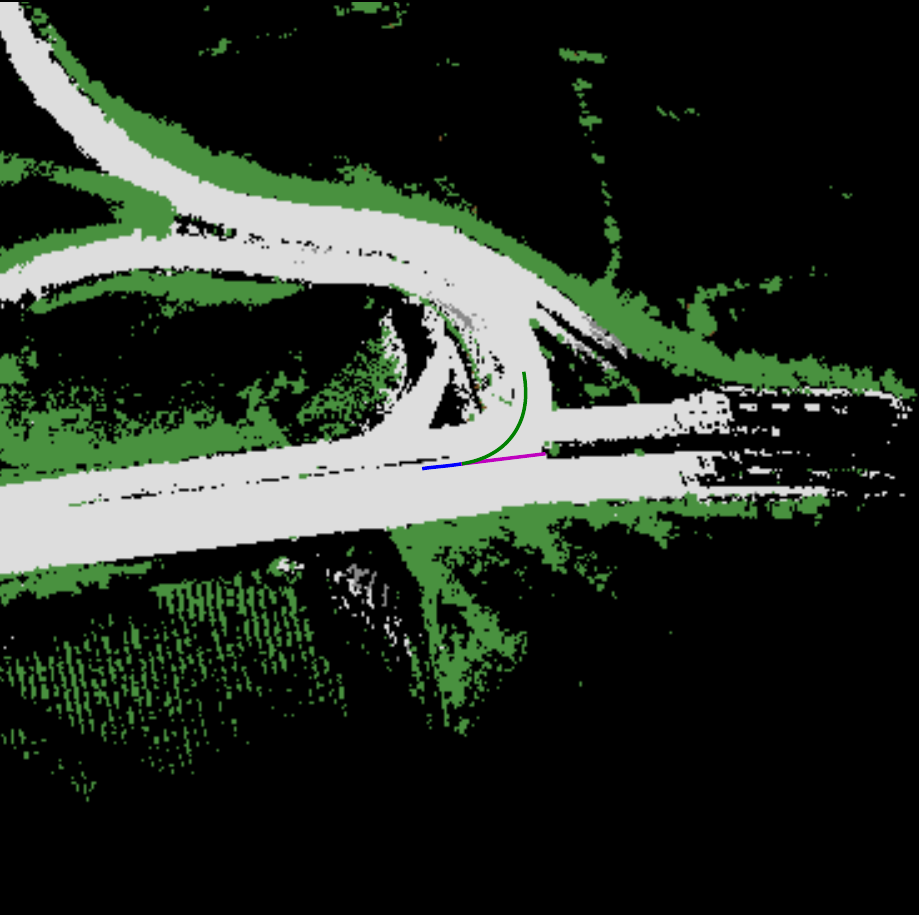} &
		\includegraphics[width=\qualw, trim=50 105 50 90, clip]{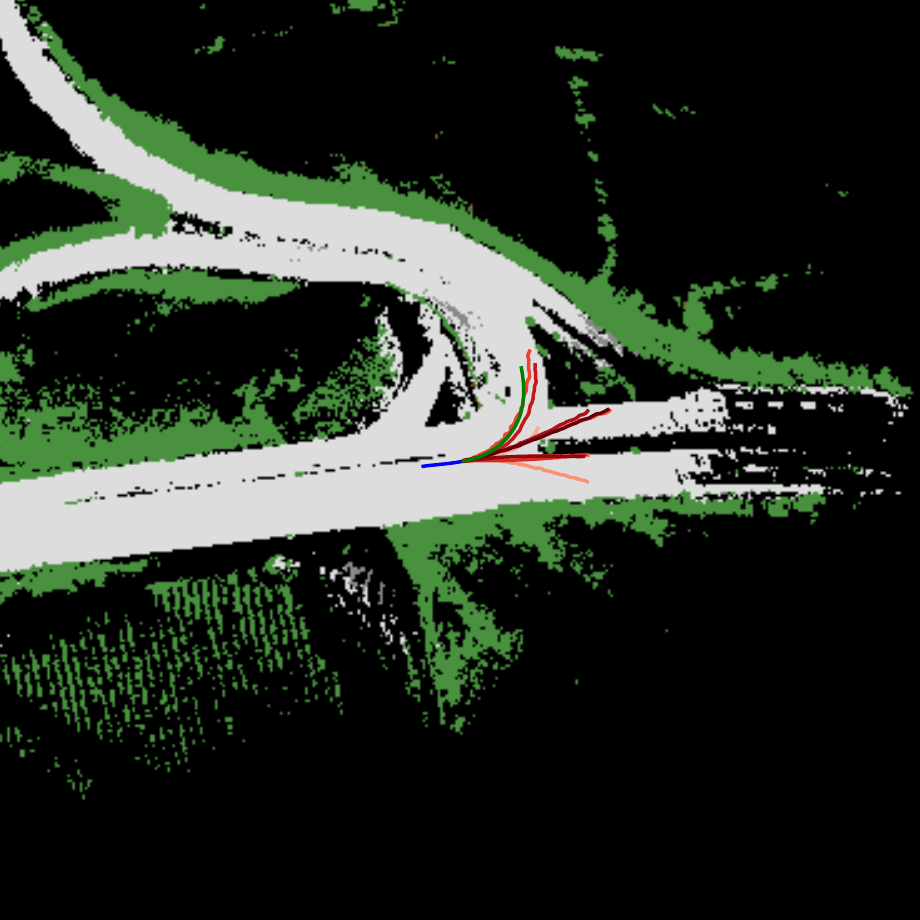}\\
		
		\includegraphics[width=\qualw, trim=50 105 50 89, clip]{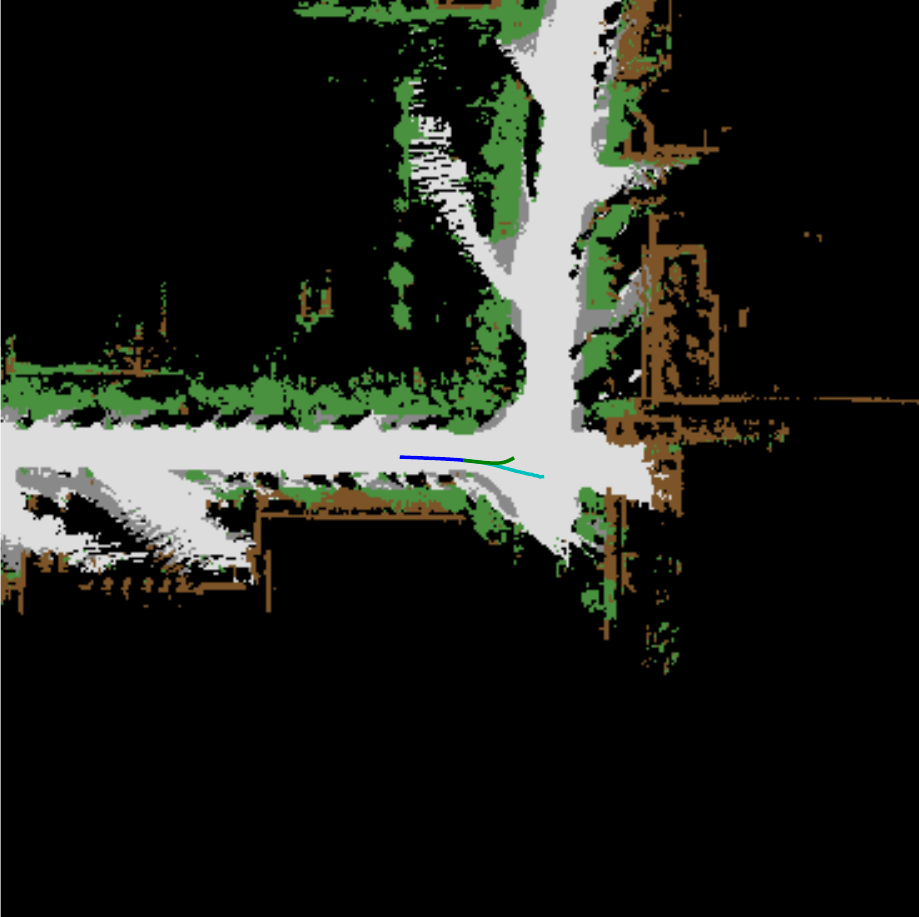} &
		\includegraphics[width=\qualw, trim=50 105 50 89, clip]{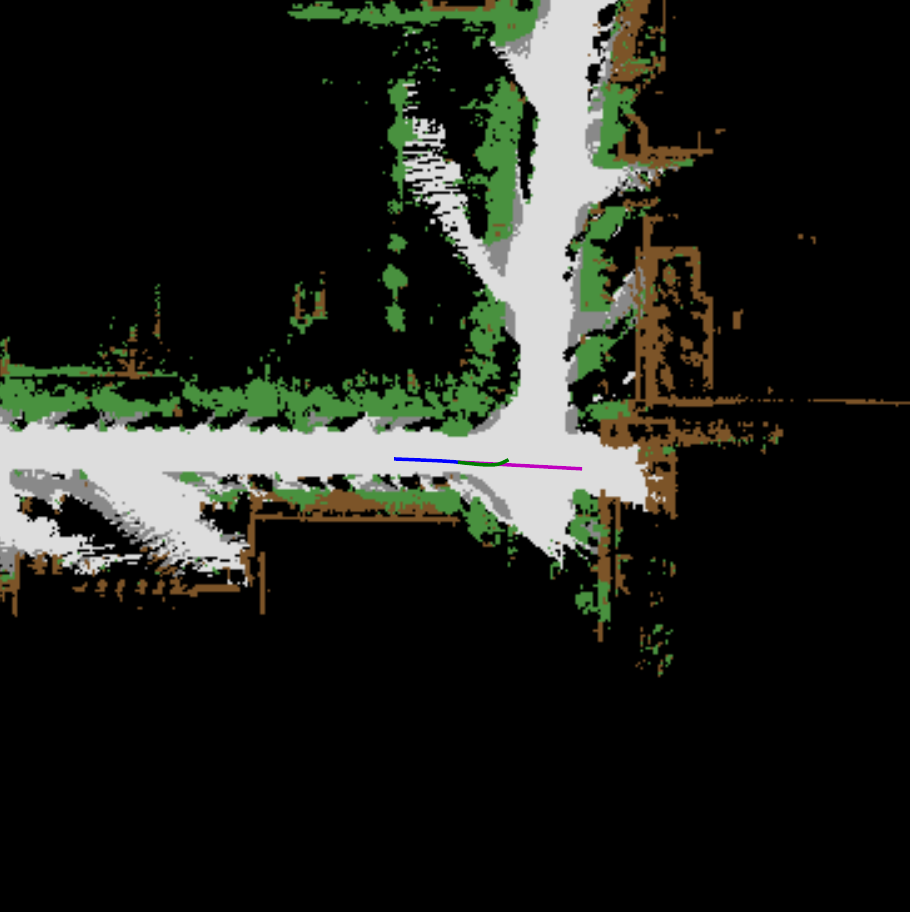} &
		\includegraphics[width=\qualw, trim=50 105 50 90, clip]{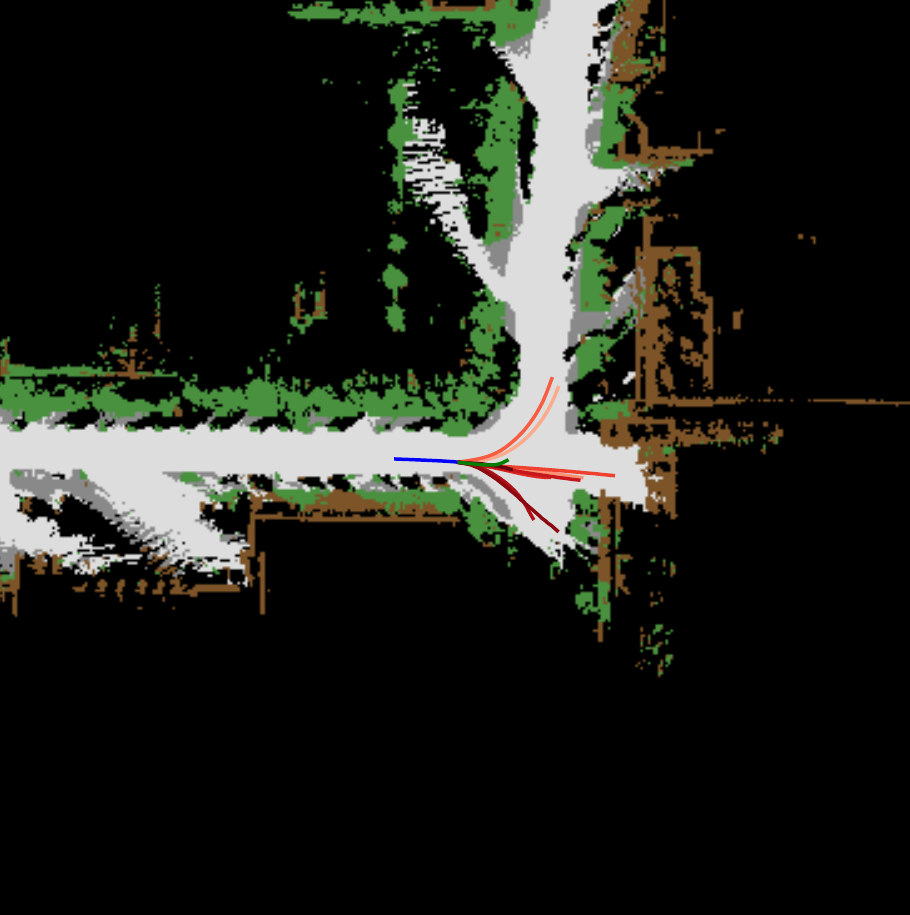}\\
		
		\includegraphics[width=\qualw, trim=50 70 0 100, clip]{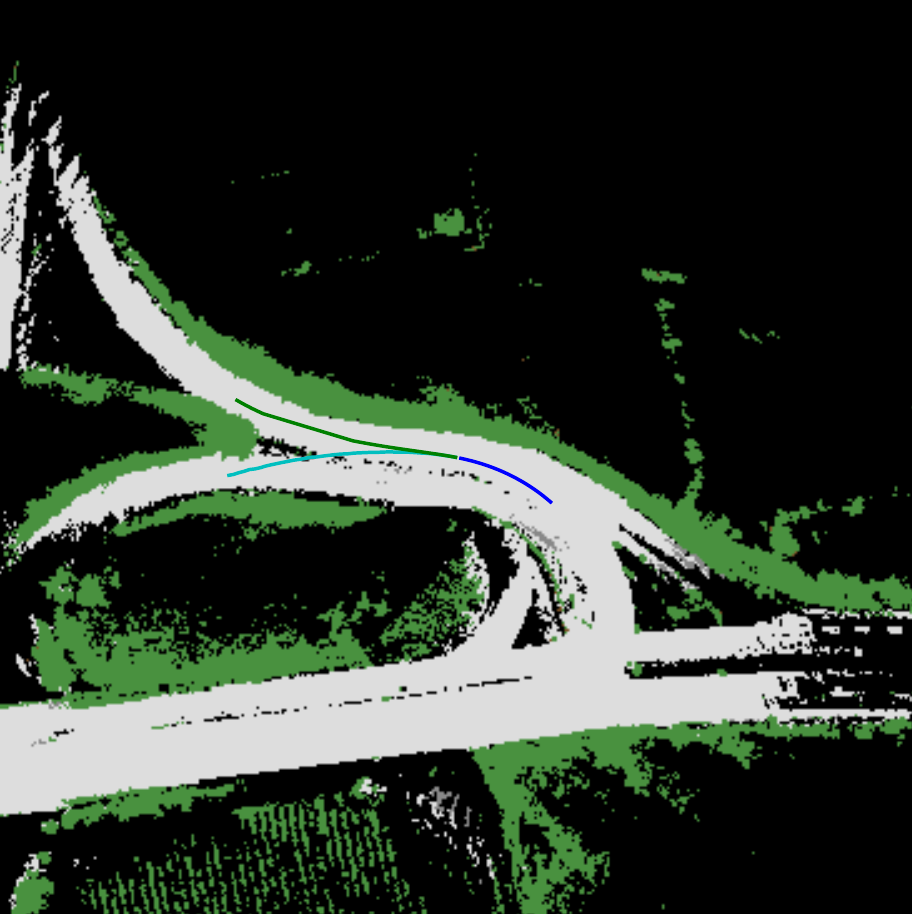} &
		\includegraphics[width=\qualw, trim=50 70 0 100, clip]{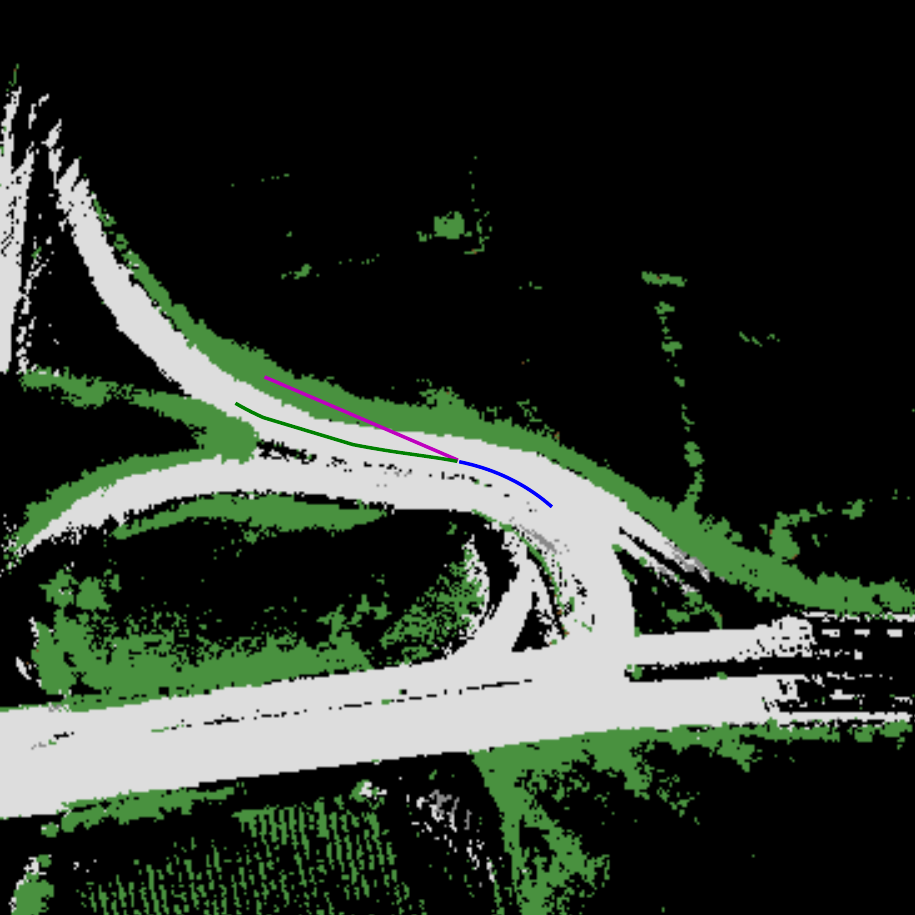} &
		\includegraphics[width=\qualw, trim=50 70 0 100, clip]{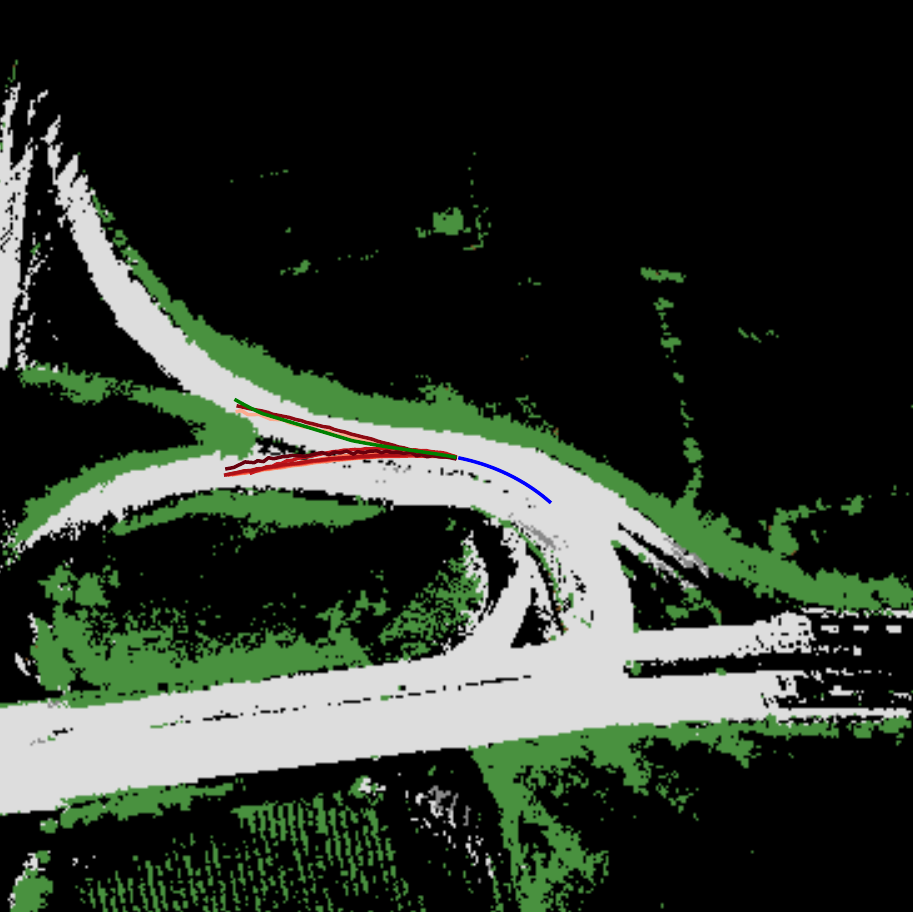}\\
		(a) Linear & (b) Kalman & (c) MANTRA\\
		
	\end{tabular}	
	\caption{MANTRA compared to Linear regression (a) and Kalman filter (b).  Methods (a),(b)  lack multi-modal capability. Past trajectories are depicted in blue, ground truth in green and future predictions are cyan (a), purple (b) and red (c). In (c) highly ranked are darker. \label{fig:qualitative}}
\end{figure*}

Following~\cite{srikanth2019infer}, we showcase the ability of our model to zero-shot transfer to other datasets. On Oxford RobotCar (Tab.~\ref{tab:oxford}) MANTRA is still able to provide satisfactory results, consistently outperforming INFER across timesteps for multimodal predictions. Analogously, on Cityscapes (Tab.~\ref{tab:cityscapes}) the model obtains a lower error than the other methods. Here we report only errors at 1s in the future, which is the maximum length of the trajectories in the dataset.

\begin{table}[t]
	\begin{center}
		\resizebox{0.7\textwidth}{!}{\begin{minipage}{\textwidth}
				\begin{tabular}{c|cccc|cccc}
					& \multicolumn{4}{c|}{ADE}       & \multicolumn{4}{c}{FDE}       \\
					Method & \multicolumn{1}{c}{1s} & \multicolumn{1}{c}{2s} & \multicolumn{1}{c}{3s} & \multicolumn{1}{c|}{4s} & \multicolumn{1}{c}{1s} & \multicolumn{1}{c}{2s} & \multicolumn{1}{c}{3s} & \multicolumn{1}{c}{4s} \\ \hline
					INFER (top 1)~\cite{srikanth2019infer}       &   1.06  &  1.35   &  1.48   &  1.68   &  1.31   &  1.71   &  1.70   &  2.56   \\
					INFER (top 5)~\cite{srikanth2019infer}       &   0.85  &  1.14   &  1.29   &  1.50   &  1.18   &  1.58   &  1.58   &  2.41   \\
					MANTRA (top 1)   &  0.55   &  0.77   &  1.01   &  1.30    &  0.60    &  1.15    &  1.82     &  2.63    \\ 
					MANTRA (top 5)  & 0.55  & 0.68      & 0.82      & 1.03     & 0.58      &  0.88    & 1.37     & 2.07    \\
					MANTRA (top 10) &  0.44     &  0.56     &  0.72     &  0.94     &  0.48     & 0.73    & 1.33 & 1.98    \\
					MANTRA (top 20) &  \textbf{0.31 }  &  \textbf{0.43 }    & \textbf{0.59}      & \textbf{0.83 }     &  \textbf{0.35 }    &  \textbf{0.61}     &   \textbf{1.24}    & \textbf{1.96 }     \\ \hline
				\end{tabular}
		\end{minipage}}
		\caption{Results on the Oxford RobotCar dataset. }
		\label{tab:oxford}
	\end{center}
\end{table}


\begin{table}[]
	\vspace{-15px}
	\begin{center}
		\resizebox{0.55\columnwidth}{!}{
			\begin{tabular}{c|c|c}
				Method   &      ADE   &    FDE    \\ \hline
				Conv-LSTM (top 1)~\cite{srikanth2019infer}   & 1.50          & - \\
				Conv-LSTM (top 3)~\cite{srikanth2019infer}   & 1.36          & - \\
				Conv-LSTM (top 5)~\cite{srikanth2019infer}   & 1.28          & - \\
				
				INFER (top 1)~\cite{srikanth2019infer}       & 1.11          & 1.59 \\
				INFER (top 3)~\cite{srikanth2019infer}       & 0.99          & 1.45 \\
				INFER (top 5)~\cite{srikanth2019infer}       & 0.91          & 1.38 \\
				
				MANTRA (top 1)                               & 0.81 & 1.42  \\ 
				MANTRA (top 3)                               & 0.66 & 1.15  \\ 
				MANTRA (top 5)                               & 0.60 & 1.00   \\
				MANTRA (top 10)                               & 0.54  &  0.86  \\
				MANTRA (top 20)                               & \textbf{0.49}  & \textbf{0.79}   \\ \hline
			\end{tabular}	
		}
		\caption{\vspace{-25px} Results on the Cityscapes dataset at 1s in the future. }
		\label{tab:cityscapes}
	\end{center}
\end{table}


\subsection{Incremental setting}
\begin{figure}
	\centering
	\includegraphics[width=\linewidth]{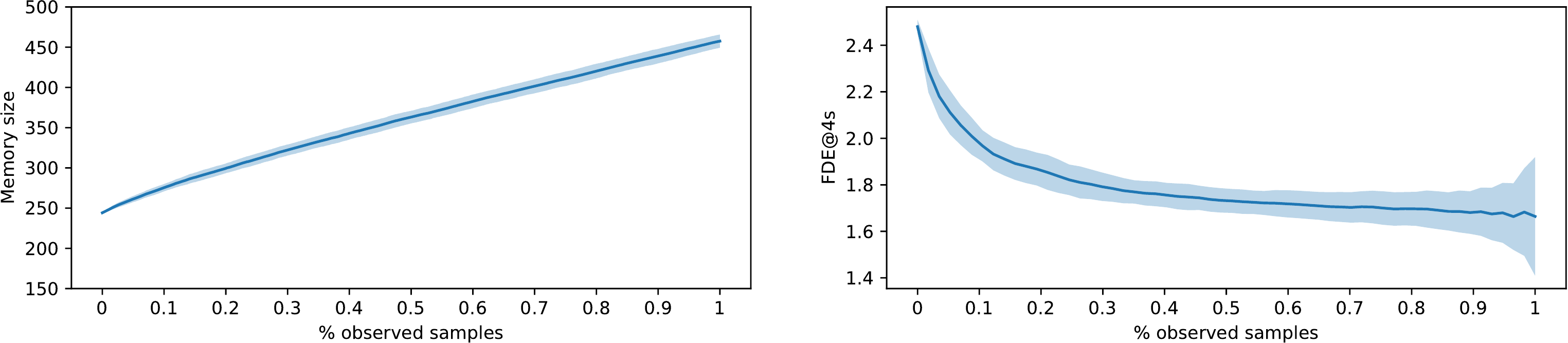}	
	\caption{Online setting. Mean and variance of memory growth (left) and error rate (right) averaged over 100 runs, increasing the observed samples.}
	\label{fig:online}
	\vspace{-10pt}
\end{figure}

Differently from prior work on trajectory prediction, MANTRA is able to improve its capabilities online, i.e. observing other agents' behaviors while driving. We simulate an online scenario on KITTI, iteratively removing a small set of 50 trajectories from the test set, presenting them to the memory  controller. The controller incorporates novel patterns according to $P(w)$. At each iteration we test the predictor on the remaining test set. 
In Fig.~\ref{fig:online} memory growth and test error are shown for MANTRA with K=5 multiple futures. Similar behaviors can be observed varying K. Interestingly, the memory size slowly grows while the error keeps decreasing. Note that the memory stores only the 16\% of the newly seen examples. To cope with the error variance increase when the remaining set of samples decreases in size we average results over 100 runs.

\section{Model Analysis}

In the following we perform ablation studies aimed at highlighting the importance of components in our model. We thoroughly investigate how the model organizes memory by checking what gets written and how it gets decoded.

\vspace{-10pt}
\paragraph{Ablation Studies}
We investigate modifications of MANTRA reporting results in Tab.~\ref{tab:ablation} on KITTI. We test the following: (i) without refinement; (ii) without decoder, i.e. reading from memory using encodings but just copying the correspondent future coordinates; (iii) without rotation invariance, i.e. using trajectories with random rotations; (iv) without memory controller, i.e. adding all training samples in memory; (v) without encoder-decoder, i.e. a Nearest Neighbor between past trajectory coordinates copying the future of the closest sample in coordinate space.
On the one hand, when the memory is filled with all training samples instead of selecting them with a controller, the error drastically increases; on the other hand even worse results are obtained when the samples are not encoded and decoded with the recurrent GRU layers. Even removing just the decoder lowers the precision of predictions considerably. This should not come as a surprise, since the decoder has the important role of adapting the suggested future from memory to the current sample, making it coherent with its past. 
Surprisingly enough instead, the refinement module does not play a very important role in the reconstruction, suggesting that the originally generated trajectories are already precise. Rotation invariance proves to be very relevant in moderating memory size and improving accuracy. By adding rotation invariance to training we lower the memory size from the 25.2\% to the 2.2\% of the observed training set. 

\begin{table}[]
	\begin{center}
		\resizebox{\columnwidth}{!}{
			\begin{tabular}{l|cccc|cccc|r}
				& \multicolumn{4}{c|}{ADE}   & \multicolumn{4}{c|}{FDE}  &  \\
				Method & 1s   & 2s   & 3s   & 4s   & 1s   & 2s   & 3s   & 4s & ~ Memory Size\\ \hline
				MANTRA (top 5)  & \textbf{0.17}  & \textbf{0.36}  & \textbf{0.61}      &  \textbf{0.94}     & \textbf{0.30}   & \textbf{0.75}   & \textbf{1.43}   & \textbf{2.48}  &  190~~ (2.2 \%)  \\
				MANTRA w/o ref. & 0.18 & 0.39 & 0.67 & 1.04 & 0.33 & 0.85 & 1.59 & 2.65 & 190~~ (2.2 \%) \\
				MANTRA w/o dec.             & 0.25 & 0.46 & 0.76 & 1.18 & 0.42 & 0.91 & 1.75 & 3.12 & 190~~ (2.2 \%) \\
				MANTRA w/o rot. inv.        & 0.25  & 0.51 & 0.88 & 1.38 & 0.45 & 1.09 & 2.10 & 3.58 & 2170 (25.2 \%)  \\
				MANTRA w/o ctrl.    		  & 0.20  & 0.45 & 0.82 & 1.34 & 0.37 & 1.02 & 2.07 & 3.64 & 8613~ (100 \%) \\
				MANTRA w/o enc-dec.            & 0.24 & 0.58 & 1.08 & 1.75 & 0.47 & 1.36 & 2.74 & 4.68 & 8613~ (100 \%) \\ 
				\hline
			\end{tabular}
		}
		\caption{Ablation study of the full method against variants without specific components: decoder, refinement, rotation invariance, trained controller, encoder-decoder. Errors are at K=5. Memory size is shown as number of samples and \% of the training set.}
		\label{tab:ablation}
	\end{center}
\end{table}

\begin{figure}
	\vspace{-10px}
	\centering
	\includegraphics[width=0.8\columnwidth]{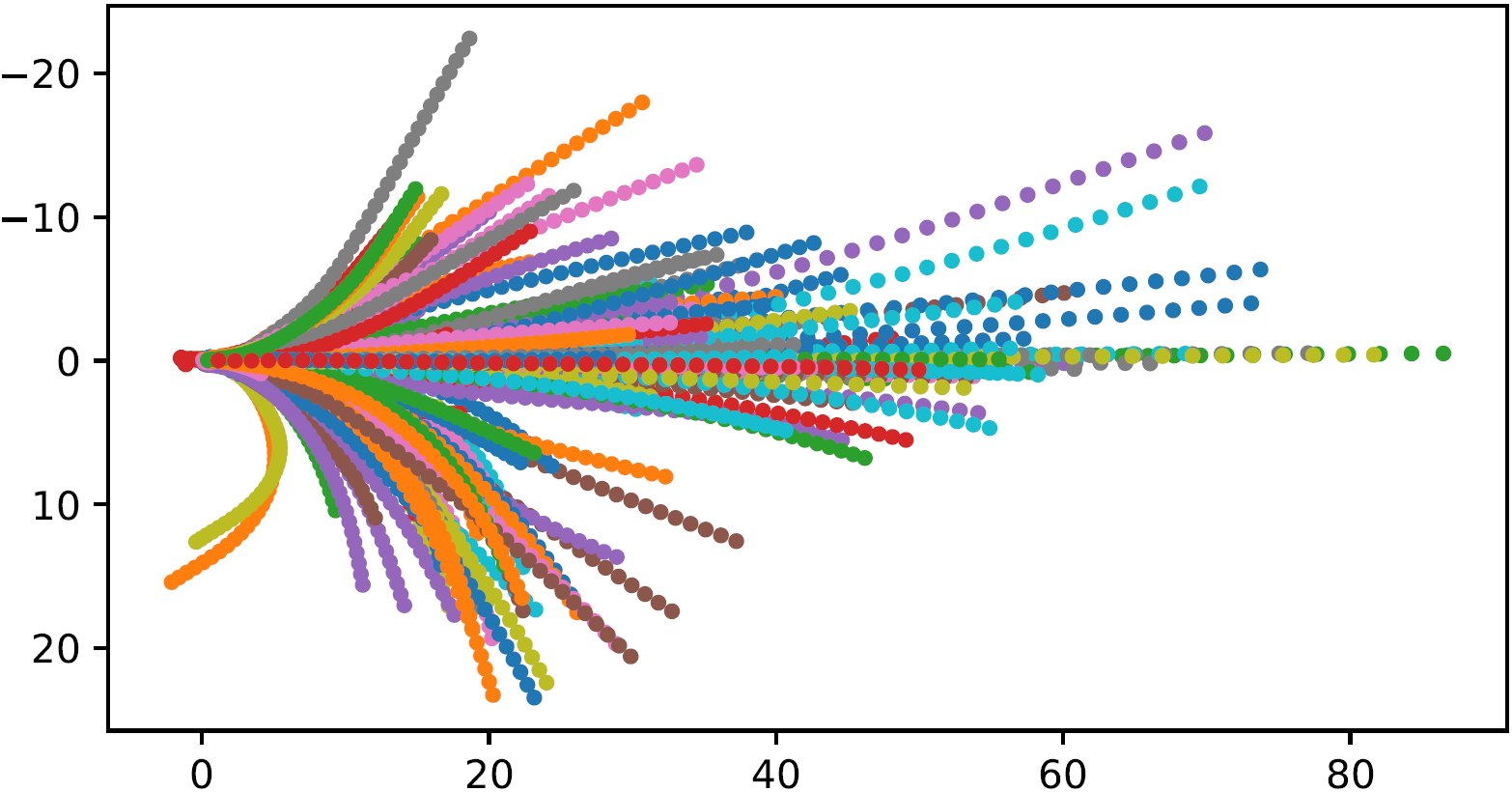}
	\caption{Decoded trajectories from memory. \label{fig:all_memory}}
	\vspace{-15px}
\end{figure}

\paragraph{Memory inspection}
To understand what the model is learning, we inspect what the controller stores in memory. We take each sample and plot its decoded future to depict a snapshot of every sample in memory. Fig.~\ref{fig:all_memory} shows all samples from a memory filled for K=5 predictions.

\begin{figure}[t]
	\centering	
	\fbox{\includegraphics[width=0.45\columnwidth]{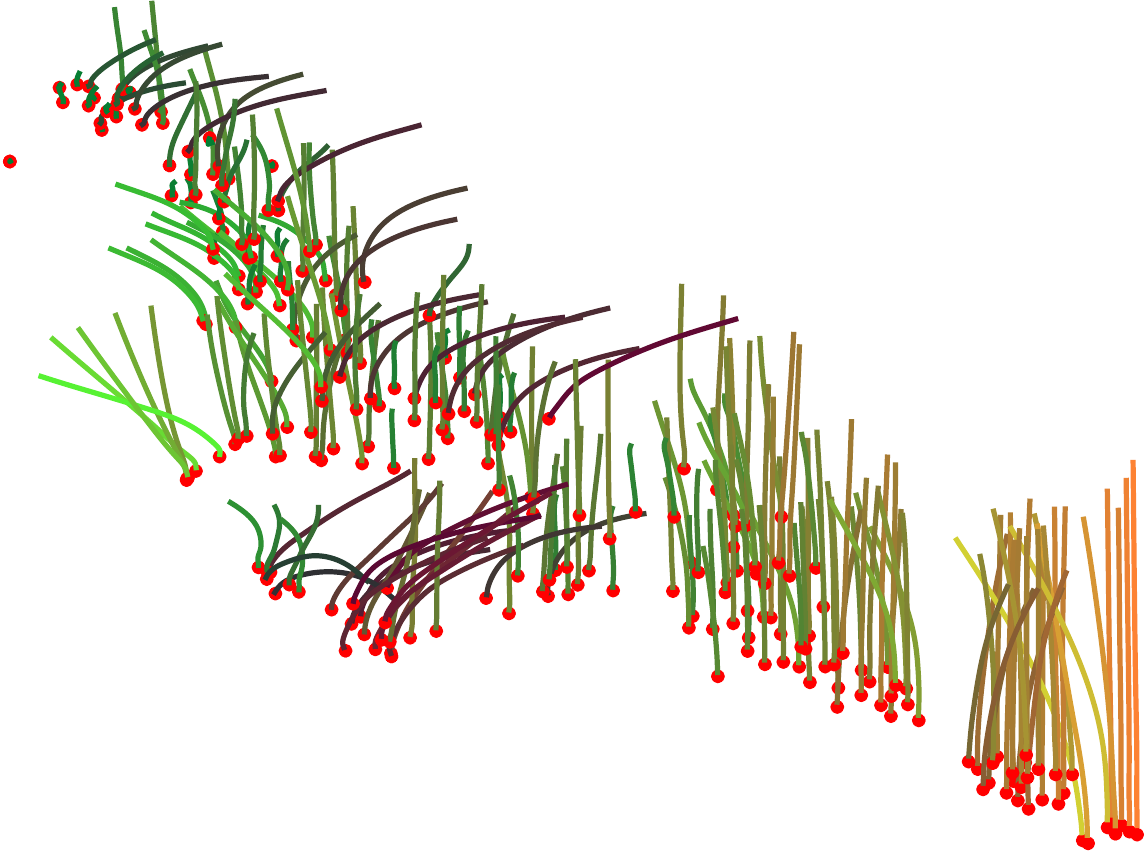}}\hspace{2pt}
	\fbox{\includegraphics[width=0.45\columnwidth]{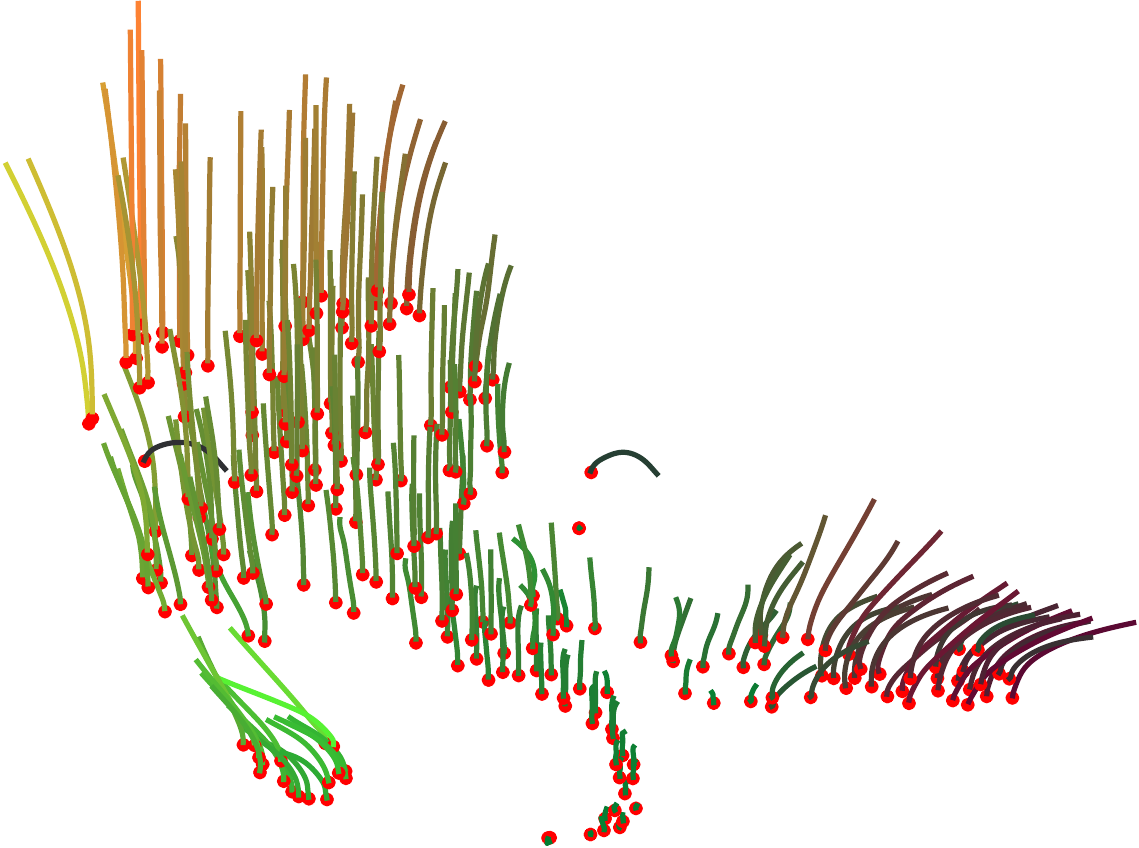}}
	\caption{T-SNE representations of past (left) and future (right) encodings stored in memory. Each point in the embedding space is shown along with the decoded trajectory. Trajectories are color coded by orientation (green tones) and speed (red tones).}
	\label{fig:tsne}
\end{figure}

In Fig.~\ref{fig:tsne} we plot T-SNE projections~\cite{maaten2008visualizing} of past and future encodings in memory, as points. On the left we plot past embeddings, while on the right we report future embeddings. For each projected sample we show future trajectories generated by the decoder, displayed starting from T-SNE points. All trajectories in the image have an upward trend due to the rotation invariance we introduce for storing samples. Similar trajectories are clustered together, indicating that the encoders are learning a manifold where similar patterns are close. Observing the T-SNE of past encodings, the multimodal nature of the problem emerges. In fact, the space appears to be organized mostly by trajectory speed and for each point several possible future directions are present. When trajectories have lower speed, futures are free to span over many possible directions, while when trajectories have higher speed, the futures vary more in length rather than curvature.

\vspace{-10pt}
\paragraph{Decoder Analysis}
We inspect the behavior of the decoder and the influence that different pasts have on future reconstructions. Encoder and decoder are jointly trained, but differently from standard autoencoders, only part of the input is reconstructed, i.e. the future. The past has the important role of conditioning the reconstruction so that we can generalize to unseen examples. In Fig.~\ref{fig:past_importance} we show several reconstructions of the same future, changing only the past encoding and keeping fixed the future one. The reconstructions of the original past yields a precise reconstruction. By changing the past by shortening it or stretching it, i.e. changing the velocity, the reconstruction gets accelerated or decelerated, affecting curvature. As a control experiment we also use a vector of zeros or a random embedding. In both cases the generated trajectories are very imprecise but still follow approximately the original trend. These tests justify using the decoder feeding a combination of encodings belonging to different samples, as we do at test time. In fact the generated trajectories are new compared to the samples in memory and they adapt to the current observation.

\newcommand{\pad}{3pt}
\newcommand{\figw}{0.18\columnwidth}
\begin{figure}[t]
	\begin{tabular}{c@{\hspace{\pad}}c@{\hspace{\pad}}c@{\hspace{\pad}}c@{\hspace{\pad}}c@{\hspace{\pad}}}
		\centering	
		\includegraphics[width=\figw]{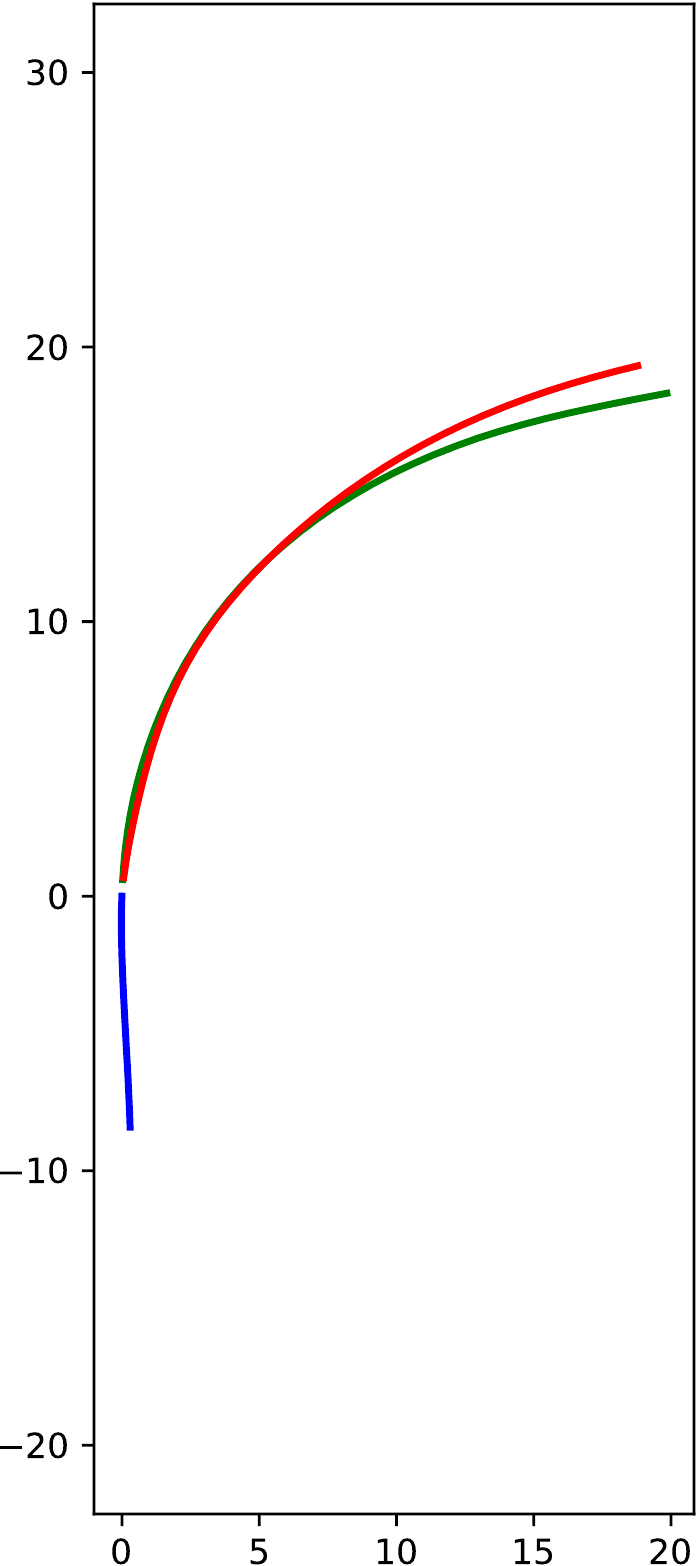}&
		\includegraphics[width=\figw]{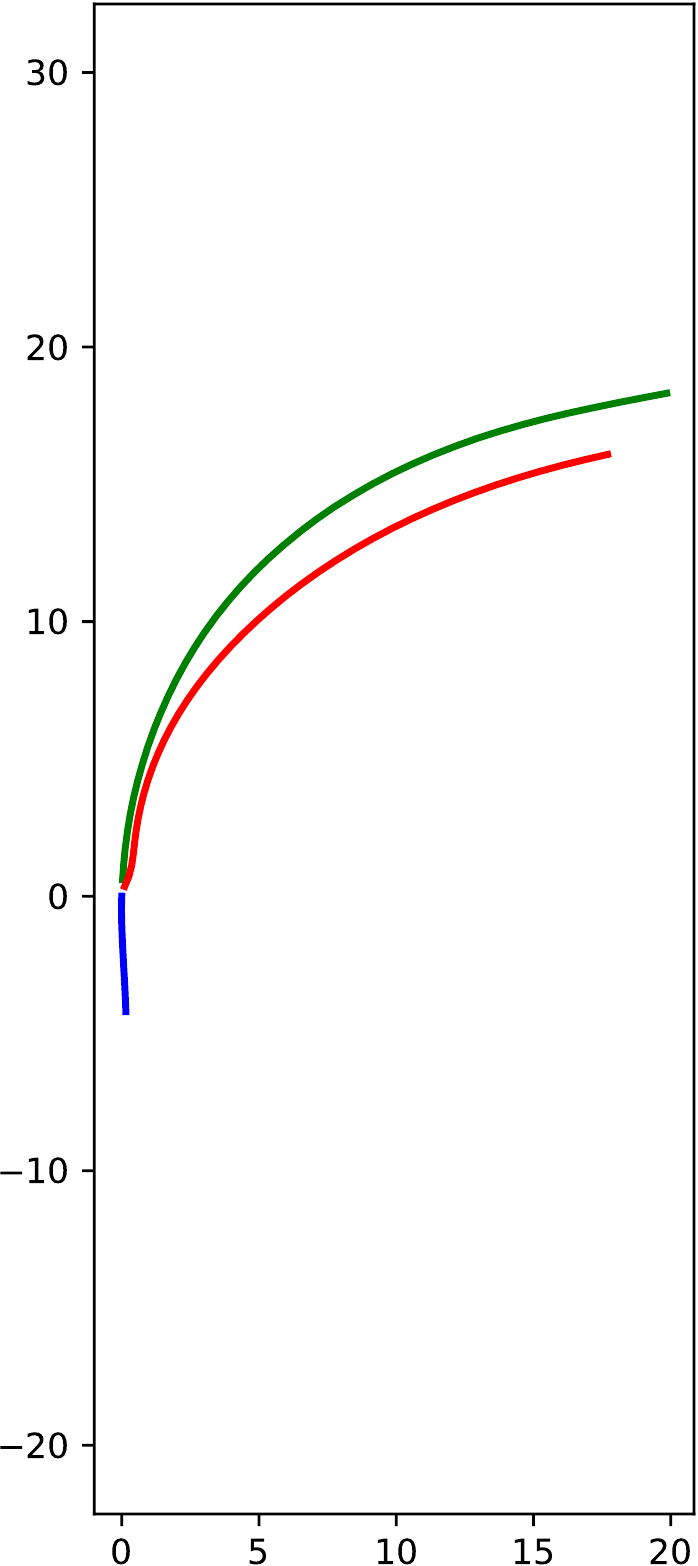}&
		\includegraphics[width=0.181\columnwidth]{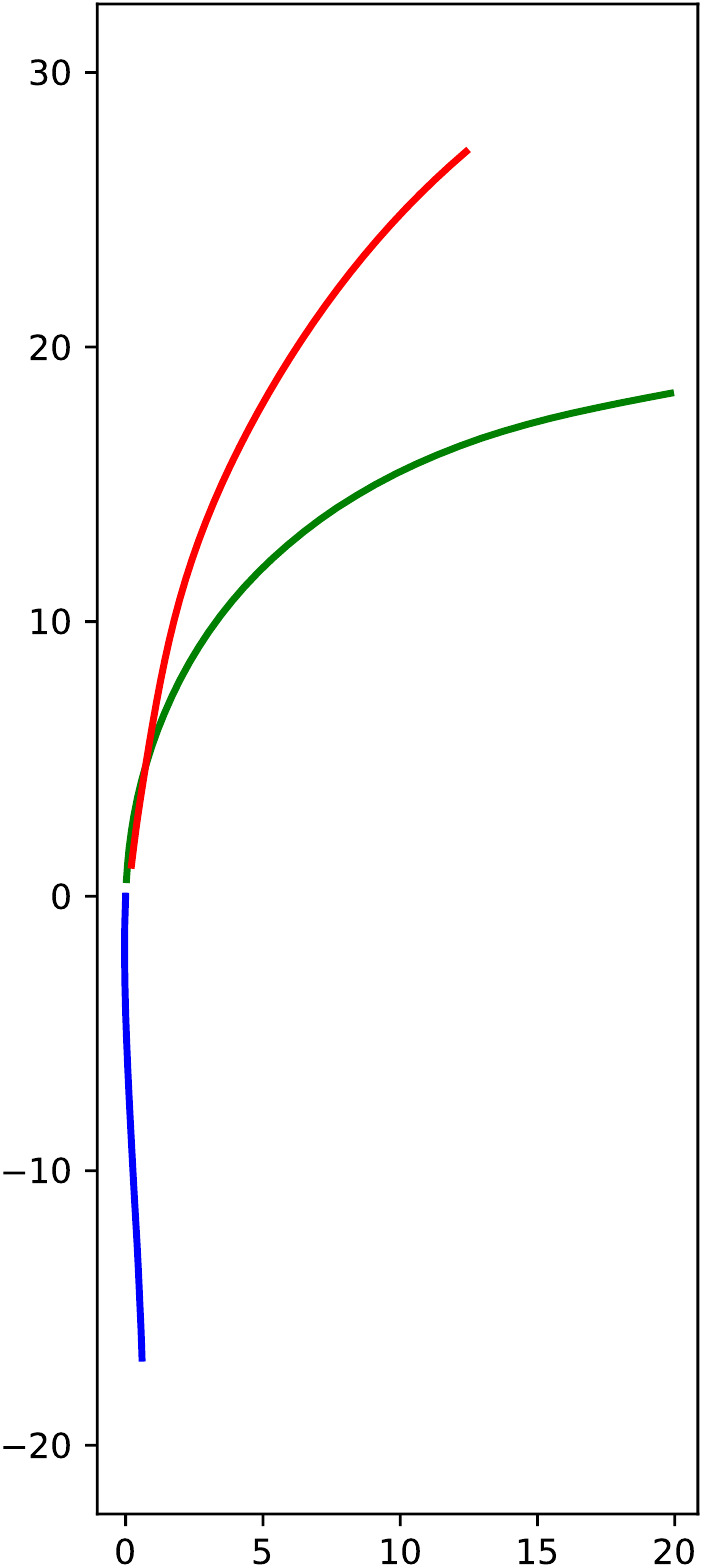}&
		\includegraphics[width=\figw]{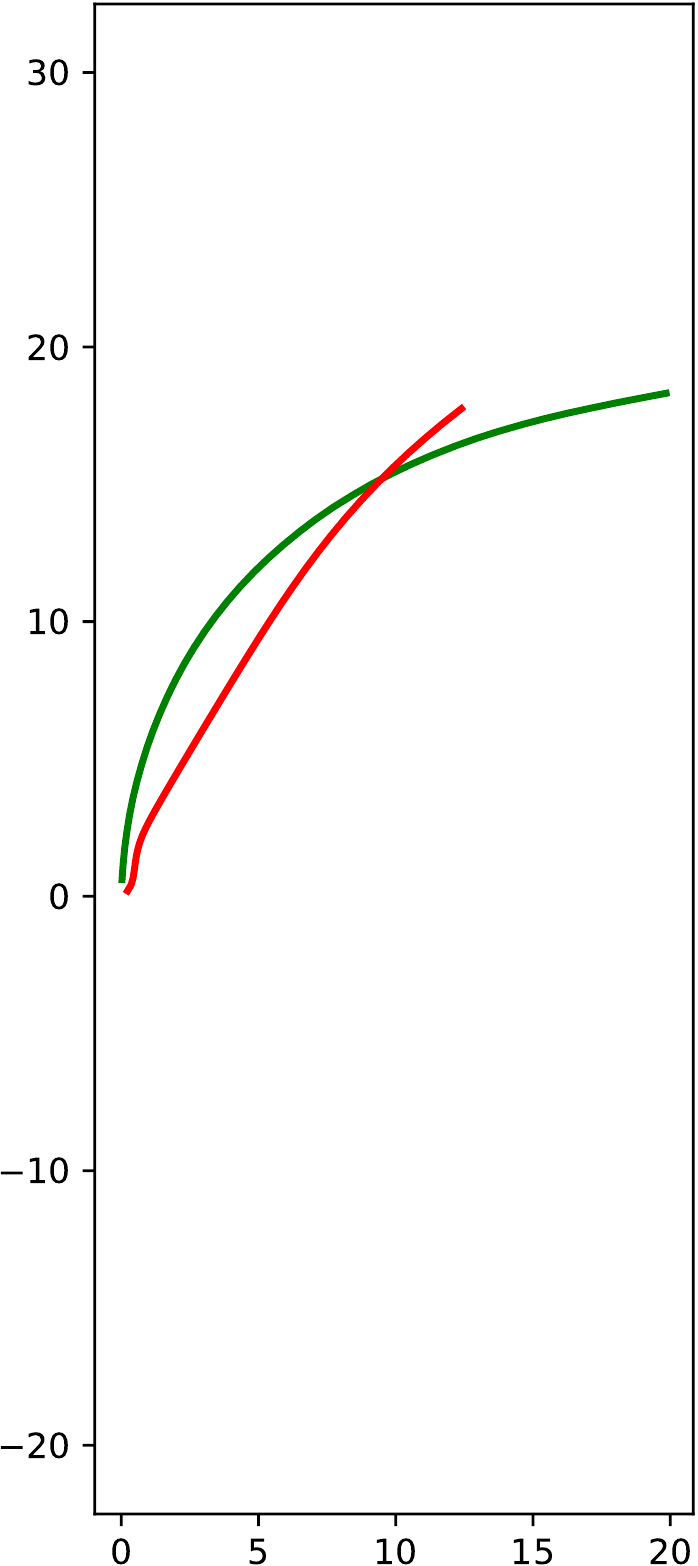}&
		\includegraphics[width=0.183\columnwidth]{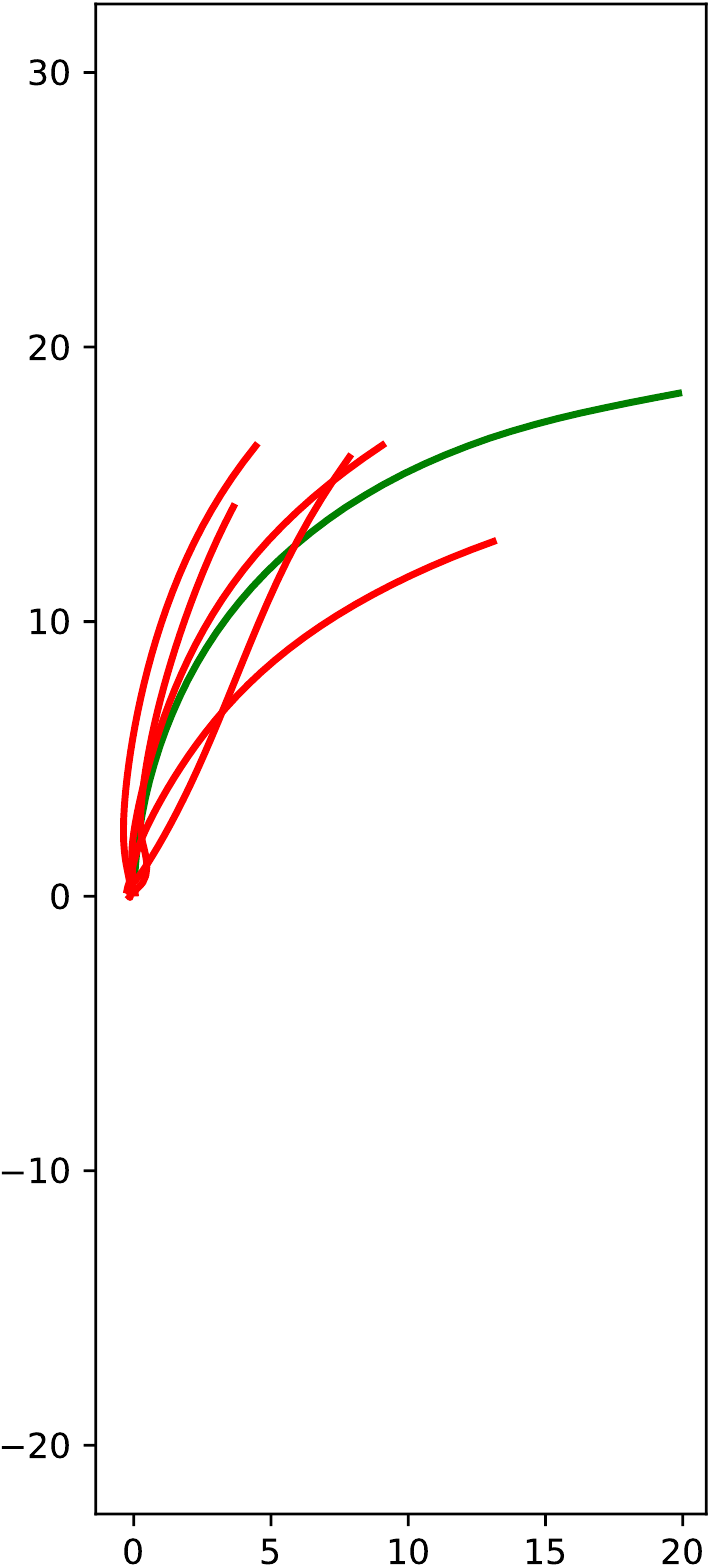}\\
		\hspace{6pt}(a)&\hspace{6pt}(b)&\hspace{6pt}(c)&\hspace{6pt}(d)&\hspace{6pt}(e)
	\end{tabular}
	\caption{Influence of past in the decoder. (a) observed past; (b) slower past; (c) faster past; (d) past embedding zeroed; (e) multiple randomized past embeddings.	Blue: past trajectory used for decoding. Red: future reconstruction. Green: original future.\label{fig:past_importance}}
	\vspace{-10pt}
\end{figure}


\vspace{-5pt}
\section{Conclusions}
We propose MANTRA, the first Memory Augmented Neural TRAjectory prediction framework. Our method, based on an associative memory, can natively grasp the inherently multi-modal nature of the future trajectory prediction problem, yielding state of the art results on three traffic datasets. Moreover, we show that the memory is able to ingest novel samples lowering the error on unseen data. 

\paragraph{Acknowledgments}
\small
We thank NVIDIA for donating a Titan Xp GPU. This work is partially founded by IMRA Europe S.A.S.

{\small
	\bibliographystyle{ieee_fullname}
	\bibliography{egbib}
}


\begin{table*}[htb!]
	\centering
	\begin{tabular}{l|r|r|r|r|r|c}
		\multicolumn{1}{c|}{\textbf{Method}} & \multicolumn{1}{c|}{\textbf{ETH}}        & \multicolumn{1}{c|}{\textbf{HOTEL}} & \multicolumn{1}{c|}{\textbf{UNIV}} & \multicolumn{1}{c|}{\textbf{ZARA1}} & \multicolumn{1}{c|}{\textbf{ZARA2}} & 
		\multicolumn{1}{c}{\textbf{AVERAGE}}\\ \hlineB{3}
		Social-GAN \cite{gupta2018social}                                  & 0.81/1.52                      & 0.72/1.61                 & 0.60/1.26                & 0.34/0.69                 & 0.42/0.84 & 0.58/1.18                 \\
		SoPhie \cite{sadeghian2019sophie}                                 & 0.70/1.43                      & 0.76/1.67                 & 0.54/1.24                & 0.30/0.63                 & 0.38/0.78 & 0.54/1.15                 \\
		CGNS \cite{li2019cgns}                                   & 0.62/1.40                       & 0.70/0.93                  & 0.48/1.22                & 0.32/0.59                 & 0.35/0.71   & 0.49/0.97               \\
		S-BiGAT \cite{Kosaraju2019BIGAT}                                & 0.69/1.29                      & 0.49/1.01                 & 0.55/1.32                & 0.30/0.62                  & 0.36/0.75 &  0.48/1.00           \\
		MATF \cite{zhao2019multi}                                   & 1.01/1.75                      & 0.43/0.80                  & 0.44/0.91                & 0.26/0.45                 & 0.26/0.57      & 0.48/0.90              \\
		GOAL-GAN \cite{Dendorfer_2020_ACCV}                               & 0.59/1.18                      & 0.19/0.35                 & 0.60/1.19                & 0.43/0.87                 & 0.32/0.65  & 0.43/0.85                \\
		Transformer \cite{giuliari2020transformer}                            & 0.61/1.12                      & 0.18/0.30                 & 0.35/0.65                & 0.22/0.38                 & 0.17/0.32 & 0.31/0.55             \\
		PECNet \cite{mangalam2020not}                                 & 0.54/0.87 & 0.18/0.24                 & 0.35/0.60                & 0.22/0.39                 & 0.17/0.30  & 0.29/0.48                \\
		Trajectron++ \cite{salzmann2020trajectron++}                           & \textbf{0.39}/\textbf{0.83}                      & \textbf{0.12}/\textbf{0.19}                & \textbf{0.22}/\textbf{0.43}                & \textbf{0.17}/\textbf{0.32}                & \textbf{0.12}/\textbf{0.25} & \textbf{0.20}/\textbf{0.40}                \\ \hline
		\textbf{MANTRA}                                  & 0.48/0.88                      & 0.17/0.33                 & 0.37/0.81                & 0.27/0.58                 & 0.30/0.67  &  0.32/0.65               \\ 
	\end{tabular}
	\caption{
		\label{tab:ethucy}Results on the ETH/UCY datasets. Each model generates K=20 multiple predictions. Errors are expressed in meters.}

\end{table*}

\FloatBarrier

{\large \textbf{Appendix: Pedestrian Trajectory Prediction}} \medskip

We analyze the capabilities of the method to predict trajectories of pedestrians, which usually require an explicit modeling of social behaviors.

We exploit the most used datasets in literature for social trajectory prediction, namely ETH/UCY~\cite{pellegrini2010eth, lerner2007ucy} and Stanford Drones Dataset (SDD)~\cite{Robicquet2016sdd}.

ETH~\cite{pellegrini2010eth} and UCY~\cite{lerner2007ucy} are top-view datasets of pedestrians, with trajectories expressed in meters at 2.5Hz. The dataset has 5 different scenarios. In particular, ETH contains two scenarios (ETH, HOTEL) and UCY contains three additional ones (UNIV, ZARA1, ZARA2). In total there are 1536 unique pedestrians exhibiting social interactions such as group actions, collision avoidance and crossing trajectories.
We use a leave-one-out strategy, following the state of the art.

SDD~\cite{Robicquet2016sdd} is a dataset of agents roaming in a university campus. Trajectories are acquired via a bird's eye view drone at a sampling frequency of 2.5 Hz.
We use the split of Trajnet~\cite{sadeghian2018trajnet}. The dataset contain 14k scenarios with multiple pedestrians. Trajectories are expressed in pixel coordinates.

In Tab.~\ref{tab:sdd} and Tab.~\ref{tab:sdd20} we show the results on SDD, respectively for K=5 and K=20 predictions.
In Tab.~\ref{tab:ethucy} we report results on ETH/UCY for K=20 predictions.
Several baselines are reported for both datasets.

Overall, it can be seen that MANTRA is able to compete against recent methods that explicitly train a social interaction module to predict pedestrian trajectories.
Interestingly, MANTRA outperforms baselines such as Social-GAN~\cite{gupta2018social} which has a generative model to output socially acceptable trajectories of multiple interacting pedestrians. Even compared to more recent approaches, such as goal-based method~\cite{Dendorfer_2020_ACCV, mangalam2020not}, MANTRA is able to obtain comparable results.


\newpage
\vspace{-1000px}
\begin{table}[]
	\centering
	\begin{tabular}{l|r|r}		
		\multicolumn{1}{c|}{\textbf{Method}} & \multicolumn{1}{c|}{\textbf{ADE}} & \multicolumn{1}{c}{\textbf{FDE}} \\ \hlineB{3}
				
		Social-GAN \cite{gupta2018social} & 27.25                             & 41.44                             \\
		CGNS \cite{li2019cgns}             & 15.60                              & 28.20                              \\
		SoPhie \cite{sadeghian2019sophie}  & 16.27                             & 29.38                             \\
		CF-VAE \cite{bhattacharyya2020conditional}                               & 12.60                              & 22.30                              \\
		P2TIRL \cite{deo2020trajectory}                                & 12.58                             & 22.07                             \\
		Goal-GAN \cite{Dendorfer_2020_ACCV}                              & 12.20                              & 22.10                              \\
		SimAug \cite{liang2020simaug}                                & 10.27                             & 19.71                             \\
		PECNet \cite{mangalam2020not}                                & \textbf{9.96}                              & \textbf{15.88}                             \\ \hline
		\textbf{MANTRA}                                  & 8.96                              & 17.76                             \\ 
	\end{tabular}
\caption{
	\label{tab:sdd20}Results on SDD for K=20 predictions. Errors are expressed in pixels. \vspace{-200px}}
\end{table}

\begin{table}[]
	\centering
	\begin{tabular}{l|r|r}
		\multicolumn{1}{c|}{\textbf{Method}} & \multicolumn{1}{c|}{\textbf{ADE}} & \multicolumn{1}{c}{\textbf{FDE}} \\ \hline
		DESIRE \cite{lee2017desire}       & 19.25                             & 34.05                             \\
		Ridel et al. \cite{ridel2020scene}& 14.92                             & 27.97                             \\
		PECNet \cite{mangalam2020not}     & 12.79                             & 25.98                             \\
		TNT \cite{Zhao2020TNTTT}          & \textbf{12.23}                             & \textbf{21.16}                             \\ \hline
		\textbf{MANTRA}	& 13.51                             & 27.34                             \\
	\end{tabular}
	\caption{
		\label{tab:sdd}Results on SDD for K=5 predictions. Errors are expressed in pixels.}
\end{table}

\end{document}